\documentclass[journal]{IEEEtran}
\usepackage{tabularx}
\usepackage{tikz}
\usepackage{hyperref}
\usepackage{gensymb}
\usepackage{soul}
\usepackage{multirow}
\usepackage{cite}
 
\usepackage{graphicx}
\usepackage{float}
\usepackage{subfigure}
\usepackage{graphicx}
\usepackage{textcomp, gensymb}
\usepackage{amsmath,amssymb,amsfonts}
\usepackage{pifont}
\newcommand{\cmark}{\ding{51}}%
\begin{document}
%
\title{Distributed Machine Learning for UAV Swarms:  Computing, Sensing, and Semantics
}
%
%
%

\author{Yahao~Ding,~\IEEEmembership{Student Member,~IEEE,}
        Zhaohui~Yang,~\IEEEmembership{Member,~IEEE,}
        Quoc-Viet~Pham,~\IEEEmembership{Member,~IEEE}
        
       Zhaoyang~Zhang,~\IEEEmembership{Senior Member,~IEEE,}
        Mohammad~Shikh-Bahaei,~\IEEEmembership{Senior Member,~IEEE,}
\thanks{Y. Ding and M. Shikh-Bahaei are with Engineering Department, King's College London, Emails: \{yahao.ding, m.sbahaei
\}@kcl.ac.uk.}
\thanks{Z. Yang  and Z. Zhang are with the College of Information Science and Electronic Engineering, Zhejiang University, Hangzhou, China, and also with Zhejiang Provincial Key Laboratory of Info. Proc., Commun. \& Netw. (IPCAN), Hangzhou, China, Emails: \{yang\_zhaohui, ning\_ming
\}@zju.edu.cn.}
\thanks{Quoc-Viet Pham is with the Research Institute of Computer, Information and
Communication, Pusan National University, Busan 46241, South Korea , E-mail: vietpq90@gmail.com.}

}

\maketitle



%
\IEEEpeerreviewmaketitle

\begin{abstract}
Unmanned aerial vehicle (UAV) swarms are considered as a promising technique for next-generation communication networks due to their flexibility, mobility, low cost, and the ability to collaboratively and autonomously provide services. Distributed learning (DL) enables UAV swarms to intelligently provide communication services, multi-directional remote surveillance, and target tracking. In this survey, we first introduce several popular DL algorithms such as federated learning (FL), multi-agent Reinforcement Learning (MARL), distributed inference, and split learning, and present a comprehensive overview of their applications for UAV swarms, such as trajectory design, power control, wireless resource allocation, user assignment, perception, and satellite communications. Then, we present several state-of-the-art applications of UAV swarms in wireless communication systems, such us reconfigurable intelligent surface (RIS), virtual reality (VR), semantic communications, and discuss the problems and challenges that DL-enabled UAV swarms can solve in these applications. Finally, we describe open problems of using DL in UAV swarms and future research directions of DL enabled UAV swarms. In summary, this survey provides a comprehensive survey of various DL applications for UAV swarms in extensive scenarios. 
\end{abstract}

\section{Introduction}
%
%
%
%
Due to the distinctive advantages of flexibility, mobility, high degree of adaptiveness and low cost, unmanned aerial vehicles (UAVs) are considered as a promising technology for 
the next generation networks 
in the aspect of 
sensing, target tracking, data collection, and providing communication services  \cite{razi2017optimal,mousavi2018leader,hayat2016survey,peng2018unified}. 
However, due to limited energy budget, computational capacity, and
flying constraints, one UAV can only serve several users in a limited time period given a limited size of network area. Therefore, it is
 necessary to investigate the use of a team of UAVs to jointly
 serve users at the network edge. UAV swarm  is a promising
 solution where several UAVs can cooperatively divide 
 a task such as forest
 fire monitoring into several sub-tasks and each UAV completes
 a sub-task individually thus reducing the task implementation
 time and improving the reliability of completing the task. Hereinafter, we refer a team of multiple UAVs as a UAV swarms.
 Future wireless networks require UAV swarms to perform tasks intelligently and autonomously,
 such as remote monitoring and mobile edge
 computing. Moreover, since the amount of data of edge devices
 grows dramatically, it is impossible for UAV swarms to upload
 all the local data they collect to the central server for machine learning (ML)
 model training and inference. To overcome these challenges, distributed learning (DL) is needed by UAV swarms.

\subsection{Motivation of DL and UAV Swarms}
DL plays an important role for UAV swarm networks. 
In particular, DL has several key applications in UAV swarm networks. First, DL such as federated learning enables UAV swarms to find a global ML model without data exchange thus enabling UAV swarms to analyze their collected data locally and improve data privacy of each UAV. Second, DL algorithms such as multi-agent reinforcement learning enables multiple UAVs to interact with the dynamic environment so as to find the optimal scheduling policy of the complicated UAV swamrs. Third, DL algorithms such as distributed inference can allow multiple UAVs to perform swarm intelligence through exchanging limited inference information.  
Fourth, DL schemes such as split learning can allow multiple UAVs to effectively train a large ML model through splitting a large model into multiple small parts. 

\subsection{Related Works and Contributions}
The goal of this survey is to review recent works on the applications of advanced DL approaches to handle the challenges in UAV swarms as well as the use of communication techniques in UAV swarms for DL performance optimization. There have been a number of surveys and tutorials on UAV swarms, ML and DL. However, they have different emphases compared to this survey. 

In\cite{mozaffari2019tutorial,bithas2019survey, lahmeri2021artificial, carrio2017review} provided a survey of using UAVs in wireless communications and mainly introduced the use of centralized machine learning for solving UAVs related problems. 
The work in \cite{mozaffari2019tutorial} provides a comprehensive survey of the potential uses of drones in wireless communications, particularly as aerial base stations and cellular-connected users. In addition challenges of UAVs such as 3D deployment, performance analysis, and energy efficiency are discussed.
The authors in \cite{bithas2019survey} introduces the ML research for UAV communication, which is presented according to the type of application in UAV. On the contrary, the work in \cite{lahmeri2021artificial} summarizes various AI methods used in UAV network, which is surveyed according to ML methods. In each ML, the UAV applications that apply this algorithm are introduced. Similarly, the use of deep learning approaches for UAVs is presented in \cite{carrio2017review}, analyzing method performances, limitations and challenges. However, these surveys did not introduce UAV swarms and their applications for communication systems.

Meanwhile, a number of surveys and tutorials \cite{skorobogatov2020multiple,tahir2019swarms,chung2018survey,zhou2020uav, azoulay2021machine,9915455} discussed the classification, performance, management, communication technologies and  application scenarios of UAV swarms. However, they did not focus on introduce the use of DL for UAV swarms. The survey in \cite{skorobogatov2020multiple} paid attention to the multi-UAV applications in recent years, listing various multi-UAVs application scenarios, classifying multi-UAVs systems and discussing data processing and communication technologies applied in multi-UAVs. Another work in \cite{tahir2019swarms} investigated drones and drone swarms and provide a discussion of the classification, functions, flight mechanisms application field and autonomy of drone swarms. In \cite{chung2018survey}, the authors conducted a survey on the state-of-the-art algorithms which enable members of the swarm to communicate, allocate assignments and collaborate to accomplish the tasks of the swarm. In particular, the work in \cite{chung2018survey} discussed the research related to trajectory generation, adversarial control, distributed sensing, tasking and detection of UAV swarms are summarized. The authors provide a comprehensive study of UAV swarm intelligence in \cite{zhou2020uav}, dividing UAV intelligence technology into five layers from a layered perspective; decision-making layer, path planning layer, control layer, communication layer and application layer. The research methods for each layer and the relationships between the layers are described, and finally, the future development trend of UAV swarm intelligence is discussed. The survey in \cite{azoulay2021machine} introduces the specific ML methods applied to solve the challenge of UAV flows in terms of flock formation, mobility aspect and communication. Recently, the first review covering all aspects of UAV swarm networks in 6G is presented in \cite{9915455}, which includes security and privacy, intelligence, and energy-efficiency problems. Moreover, blockchain and some AI/ML algorithms are introduced to establish 
ultimate UAV swarm networks in \cite{9915455}. However, these investigations did not focus on studying the use of DL for UAV swarms. 

The literature \cite{brik2020federated,9739009} both mainly discussed the FL application for UAVnetworks. The use of Federated deep learning (FDL) in UAV-enabled wireless networks is investigated in \cite{brik2020federated}.
In particular, it described the conception of FDL, possible FDL applications in UAV networks and how to handle targe challenges via FDL. The survey in \cite{9739009} fuses blockchain and FL in UAV networks and proposes a taxonomy of blockchain-based FL based solutions for UAVs in B5G networks. However, these surveys did not present other DL algorithms applied to UAV networks. 
Recently, several existing surveys \cite{9562559,9733984,letaief2021edge} investigated advanced DL algorithms used in wireless networks. The work in \cite{9562559} describes the importance of DL in wireless networks and how to deploy DL frameworks in wireless edge networks effectively. It slightly mentions drones, but not swarms of drones. Other authors provide a survey of DL technologies in wireless communication in \cite{9733984}, detailing several distributed frameworks and algorithms, analyzing examples and application prospects of DL in the physical layer, media access layer control layer, and network layer. Moreover, the achievement of edge AI systems for 6G is investigated in \cite{letaief2021edge}, with a discussion of the edge training and inference methods and resource allocation and system architecture. However, these surveys do not mention UAV swarms, but the framework presented can be applied to UAV swarms in the future.

\begin{table*}[!htb]
    \centering
    \caption{Existing surveys on UAVs and ML}
    \label{tab:summary}
    \begin{tabular}{|c|c|c|c|c|c|c|}
    \hline
     Ref. & UAVs  & Swarms& ML& DL & Key contributions & Limitations\\
     \hline
     
      \cite{9562559} & & & &\cmark & \begin{tabular}{p{6cm}}A survey on DL deployed in wireless edge networks, including hardware resources ,wireless environment and resources.\end{tabular} &\begin{tabular}{p{4.5cm}}  Only a little mention of UAV, but no focus on UAV swarms.\end{tabular} \\
    \hline
     
     \cite{mozaffari2019tutorial} & \cmark &  & & &\begin{tabular}{p{6cm}}A survey on UAVs applications and challenges in wireless networks, including the investigation of  two main cases aerial BS and cellular-connected user and the discussion of  the design and optimization for UAV wireless communication system.\end{tabular}  & \begin{tabular}{p{4.5cm}}This paper does not focus on DL methods used on UAV swarms, only a few ML methods are mentioned.\end{tabular} \\
     \hline
     
    \cite{bithas2019survey} & \cmark & &\cmark & & \begin{tabular}{p{6cm}}A detail survey for ML algorithms used on UAV-based communication for improving various design and functional aspects.\end{tabular} &\begin{tabular}{p{4.5cm}}  This paper does not focus on DL used on UAV swarms.\end{tabular} \\
    \hline
    
    \cite{lahmeri2021artificial} & \cmark & & \cmark &  &  \begin{tabular}{p{6cm}} A survey on AI approaches for UAV network, such as supervised and unsupervised learning, RL and FL.\end{tabular} & \begin{tabular}{p{4.5cm}}This paper does not focus on DL used on UAV swarms, Only the FL algorithm is mentioned in the distributed algorithm.\end{tabular} \\
    \hline
    
    \cite{carrio2017review} & \cmark & & \cmark & & \begin{tabular}{p{6cm}}A brief survey on the applications of deep learning for UAVs, including the introduction of the key deep learning algorithms and the analyse the performance and limitations of deep learning for UAVs.  \end{tabular} &\begin{tabular}{p{4.5cm}} Only deep learning algorithms are considered, while the DL methods and UAV swarms are ignored.\end{tabular} \\
    \hline
    
    \cite{azoulay2021machine} & \cmark &\cmark & \cmark & 
    & \begin{tabular}{p{6cm}}A wide view of various aspects of ML that are applicable to flock management.\end{tabular} & \begin{tabular}{p{4.5cm}} Only few DL methods applied on UAVs are involved.\end{tabular} \\
    \hline

    \cite{brik2020federated} & \cmark & & &\cmark & \begin{tabular}{p{6cm}} A survey on the use of Federated deep learning in UAV-enabled wireless networks. \end{tabular} & \begin{tabular}{p{4.5cm}} This paper only focus on the use of  FL algorithm in UAV networks, while the other distributed learning algorithms and  UAV swarms scenario are ignored.\end{tabular} \\
    \hline 
    
     \cite{skorobogatov2020multiple} &\cmark &\cmark & & &\begin{tabular}{p{6cm}} A survey on the multi-UAVs system, including the classification and applications of the multi-UAVs system and the analysis of the communication technologies. \end{tabular}& \begin{tabular}{p{4.5cm}} This paper only focus on the various applications for multi-UAVs, while DL algorithms are not considered. \end{tabular}   \\
    \hline 
    
     \cite{tahir2019swarms} &\cmark & \cmark& & &\begin{tabular}{p{6cm}} A survey on the UAV swarm, with the discussion of the classification, functions, flight mechanisms application field and autonomy. \end{tabular}&\begin{tabular}{p{4.5cm}} The DL algorithms and applications of DL in UAV swarms have not been presented.\end{tabular}\\
    \hline 
    \cite{9915455} &\cmark &\cmark &\cmark & &\begin{tabular}{p{6cm}} A technical survey on the UAV swarm for network management over 6G, including security, privacy, intelligence, and energy.    \end{tabular}&\begin{tabular}{p{4.5cm}} This paper makes little mention of DL algorithm applied in UAV swarms.\end{tabular} \\ 
    \hline
    \cite{chung2018survey} & &\cmark & \cmark & & \begin{tabular}{p{6cm}} A survey for technologies and applications on aerial swarm robotics, such as task assignment, trajectory planning and platforms.\end{tabular} &\begin{tabular}{p{4.5cm}} This paper does not focus on DL used on UAV swarms \end{tabular} \\
     \hline
    \cite{zhou2020uav}& &\cmark &\cmark & &\begin{tabular}{p{6cm}} An overview of the recent advances in UAV swarm intelligence, where the functions of 5 layers of UAV swarm intelligence and challenges of future UAV swarm research are illustrated. \end{tabular}& \begin{tabular}{p{4.5cm}} This paper only focus on discusses the 5 layers of UAV swarm intelligent, only a few ML and DL methods are mentioned. \end{tabular} \\ 
    \hline

    \cite{letaief2021edge} & & &\cmark &\cmark &\begin{tabular}{p{6cm}} An overview on edge AI system for 6G, including the new design principles, resource allocation and system architecture. \end{tabular} & \begin{tabular}{p{4.5cm}} This paper only focuses on edge AI implementations, UAV is only mentioned as an application scenario and is not primarily about UAV Swarm.\end{tabular} \\
    \hline 
    
   \cite{9733984} & & &\cmark &\cmark &\begin{tabular}{p{6cm}} An overview of the DL technologies for wireless communication, including introduction of DL architectures and approaches, the analysis of the applications and challenges of DL in different network layer.\end{tabular} &\begin{tabular}{p{4.5cm}} This paper focuses only on DL technologies for wireless communication and only slightly mentions UAVs in the application cases.\end{tabular}\\
    \hline
    
    \cite{9739009} &\cmark &\cmark & &\cmark &\begin{tabular}{p{6cm}} A survey on the blockchain-FL based UAVs applications and case study in B5G, such as edge services, healthcare, agriculture and Data Dissemination.    \end{tabular}&\begin{tabular}{p{4.5cm}} This paper only focus on BC-based FL case study for UAVs, while other DL methods and the analysis of various optimizations for UAV swarms are not considered.\end{tabular} \\ 
    \hline
    
    Our paper &\cmark & \cmark &\cmark &\cmark & \begin{tabular}{p{6cm}} An comprehensive survey on DL methods for UAV swarms. In particular, we introduce 
    
    \begin{itemize}
        \item Introduction to the advantages of UAV swarms and their development prospects
        \item A detailed introduction to four advanced DL methods
        \item A comprehensive summary of DL cases in UAV swarms
        \item Introduction of 4 new applications in combination with UAV swarms
    \end{itemize}

    \end{tabular} & \\
    \hline
    \end{tabular}
    
\end{table*}



 Table~\ref{tab:summary} summarized the differences between the existing works and this work. By analyzing these surveys, the limitations prompt us to undertake a more holistic investigation into the use of DL in UAV swarms. Therefore, the key purpose of this survey is to provide a comprehensive overview on DL enabled UAV swarm networks, which lies in the discussion of the applications of DL in various optimization problems in UAV swarm networks, including trajectory optimization, wireless research allocation, sensing, caching, computing resource allocation and user association for satellites and UAVs. In our survey, we first review four advanced DL algorithms. Then, we introduce some state-of-the-art communication techniques such as RIS, semantic communications, VR that can be used for UAV swarm networks. Finally, we discuss some important research challenges and highlight interesting future directions in UAV swarms. The major contributions are highlighted as follws:

\begin{itemize}
\item We present a state-of-the-art introduction on the application of DL in UAV swarms, from introducing the benefits of UAVs and the wide range of application scenarios, putting forward why to use UAV swarms and why DL is suitable for application in UAV swarms, to recent relevant research.

\item We describe four advanced DL methods, FL, MARL, DI and SL in detail. Their basic architectures and classifications are presented, as well as a summary of their advantages and disadvantages.

\item We conduct a holistic investigation and analysis of the optimization issues applied DL in various UAV swarms applications, including trajectory, beamforming, bandwidth, time, user allocation, power control, sensing, caching, computing research allocation and user association for satellites and UAVs. Tables summarizing the key methods, applications and objections of each DL algorithm used in UAV swarms are also provided.

\item Several novel UAV swarms application scenarios such as RIS-UAVs, semantic-UAVs and VR-UAVs are referred. The applications of DL in these novel scenarios are
also discussed.

\end{itemize}

\begin{figure*}[htpb]
    \centering
    \includegraphics[width=1\textwidth]{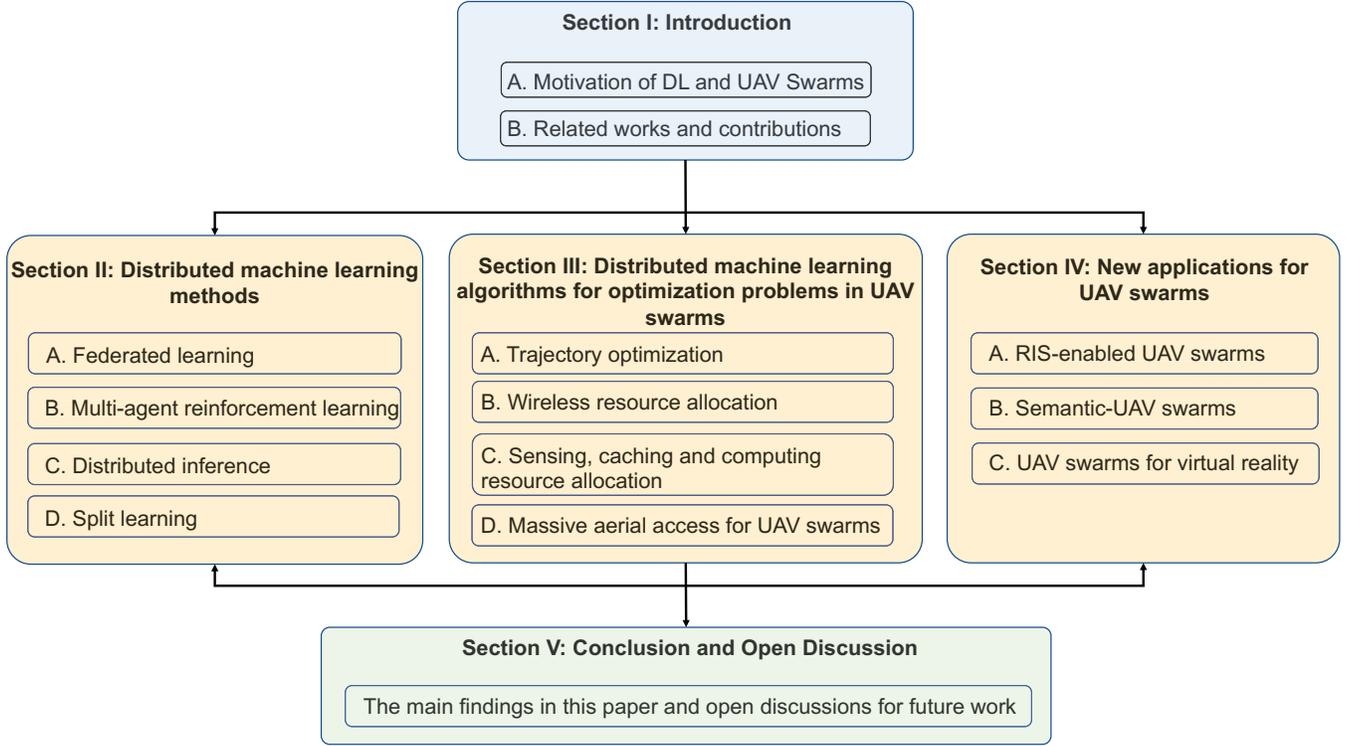}
    \caption{The overall organization of this survey.}
    \label{fig:org}
\end{figure*}

The organization of this paper is shown
in Fig.~\ref{fig:org}. Four state-of-art DL approaches including FL, MARL, DI and SL are described in Section~\ref{DL}. In each DL algorithm, we introduce its basic model and its applications in UAV swarm networks. Finally, we summarise the advantages, disadvantages and application areas of these four algorithms. Section~\ref{DS} focuses on the use of DL methods to help UAV swarms solve problems, including path optimization, wireless resource allocation, computational resource allocation, and also mentions user association for satellites and UAVs. In Section~\ref{New}, we introduce the interplay between emerging applications and UAVs, e.g., RIS, semantic communication, VR, etc., and introduce the application of DL to these new techniques. Finally, we will conclude this paper and discuss the challenging open issue that should be solved and the further research about DL-UAV swarms in Section~\ref{con}.

\section{Distributed Machine Learning Algorithms} \label{DL}
In this section, we present four state-of-the-art DL methods including federated learning, multi-agent reinforcement learning, distributed inference, and split learning. We describe their principles in detail and analyze their advantages and disadvantages.


\subsection{Federated Learning}
\begin{figure}[t]
    \centering
    \includegraphics[width=0.5\textwidth]{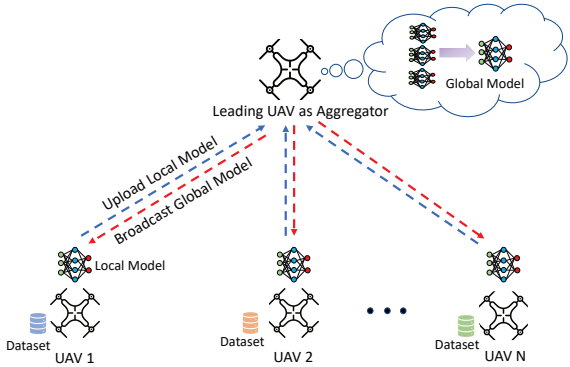}
    \caption{The system model of a UAV swarm using FL.}
    \label{fig:FL}
\end{figure}

FL is a distributed learning method proposed by Google in 2016 \cite{konevcny2016federated} \cite{konevcny2016federated1}. FL enables UAV swarms to build a machine learning model based on distributed data sets. During an FL training process, UAVs send their local trained model to a central server without sharing their raw data. In other words, in FL training, the raw data of each UAV is stored locally. 
According to the data distribution characteristics, FL can be divided into two categories: horizontal federated learning and vertical federated learning \cite{yang2019federated}. Typically FL refers to horizontal federation learning, which is essentially the union of samples, and is applicable to scenarios where participants have similar features but different samples. Vertical FL refers to the union of features, which is mainly used in scenarios in which the samples of datasets are similar between participants, but the features are different. Next, we first introduce the basic model of the typical FL algorithm.

\subsubsection{Introduction of FL}
In general, the FL training process includes the following three steps, as shown in Fig.~\ref{fig:FL}.

\textbf{(1) Training initialization:} Designating one drone as a server and other selected reliable drones as following clients. The agent first initiates a global model $W_{G}^0$ and sets up hyperparameters of training processes, e.g., the number of epochs and learning rate. The initialized global model $W_{G}^0$ is broadcast to clients in the first round.

\textbf{(2) Local training and updating:} After receiving the global model $W_{G}^j$, where $j$ denotes the current iteration index, each client $i$ uses their local data to train a local model and sends the computed local model parameters $W_{i}^j$ to the server for aggregation.


\textbf{(3) Model aggregation and download:} Once the server receives all local models, it combines them to update the global model $W_{G}^{j+1}$.Then, the updated global model parameter $W_{G}^{j+1}$ is sent back to clients for the next training round. The objective of the server is to minimize the average global loss function $L(W_{G}^j)$:
\begin{equation}
    L(W_{G}^j) = \frac{1}{M}\sum_{i=1}^{i=M}L(W_{i}^j),
\end{equation}

where $M$ denotes the number of aggregated models. $L$(·) is loss function, and the expression of the $L$(·) is application-specific. Stochastic gradient descent (SGD), as a popular algorithm, is frequently used to solve local FL loss function problems. Finally, processes 2-3 are iterated until the global loss function converges or achieves the desired accuracy. By using FL to train the global model, we find that it not only protects client UAV privacy but also reduces network overhead and latency. Moreover, FL can train different models based on different ML algorithms, according to the different applications \cite{chen2021communication}.





\subsubsection{Advantages and Disadvantages of FL}

FL is an effective distributed algorithm for protecting data privacy of clients, and  clients only need to upload the trained local model parameters to the server instead of uploading large amounts of raw data, which greatly reduces the latency and energy consumption in wireless communication networks \cite{niknam2020federated,yang2021federated}. However, FL also suffers from some drawbacks and challenges.

1. Expensive communication: Large federation networks may contain hundreds or thousands of users, and each FL training process of uploading models and broadcasting models take up a large number of communication resources. To improve communication efficiency, the number of communication rounds can be reduced or the size of information per transmission can be reduced \cite{li2020federated}.

2. Systems Heterogeneity: Due to the differences in users' hardware, battery, and network environment, there are differences in communication, computation, and storage capabilities among users. And in FL training, users are not guaranteed to participate all the time, which will have stragglers exist. In addition, it is difficult for the server to wait until all local models are accepted before aggregation because of the different processing speeds and communication abilities of users \cite{li2020federated}.

3. Data Heterogeneity: In general, the data collected by users are non-identically distributed, which leads to inconsistent features learned by the user-side model (model heterogeneity). It is possible to solve the data heterogeneity problem by building a personalized FL \cite{fallah2020personalized}.

4. Privacy Concerns: In the FL training process, there is a risk that the user's private information will be leaked, and the attacker may be the server, an external eavesdropper or a participant. Data privacy attacks are divided into three categories, membership inference, model inversion and GAN reconstruction. Membership inference attack (MIA) \cite{shokri2017membership} refers to identifying whether a given sample is in the training dataset or whether it belongs to a particular class of model representation \cite{truex2019demystifying}. Model inversion attacks \cite{fredrikson2015model} use leaked parameters to obtain the sensitive information represented by the model, for example, the gradient leakage attack is one such type of attack. GAN reconstruction attacks are similar to model inversion attacks, but it can generate a realistic sample to obtain sensitive properties of the sample \cite{hitaj2017deep}. It is suitable for attacking complex DL models. 

Model performance attacks are divided into targeted and untargeted attacks, both of which can be achieved through data poisoning \cite{biggio2012poisoning} and model poisoning. Data poisoning refers to the indirect corruption of a model by maliciously compromising training or inference data, such as label flipping and inserting back doors. Model poisoning aims to directly modify the model by manipulating the learning process. For example, directly manipulating gradients or learning rules.

These malicious attacks can recover private training data and affect the accuracy of FL training models. Recently many studies focus on defence methods for FL attacks. Differential privacy (DP) as a standard method is proposed and widely used to defend data privacy attacks by adding artificial noise at sensitive attributes \cite{van2014encyclopedia}. The aggregation algorithm (e.g., Byzantine-robust algorithm \cite{kairouz2021advances,bagdasaryan2020backdoor})  and outlier detection(e.g., reject on negative impact \cite{barreno2010security}, TRIM \cite{jagielski2018manipulating})  are primarily proposed to reduce the impact of model performance attacks \cite{jere2020taxonomy}. In \cite{su2021federated}, they considered that the additional noise brought by using DP and the unbalanced distribution of data in the local client affect the performance of the FL model, resulting in low accuracy. To solve this problem, the authors proposed a federated regularization learning model to protect data privacy in FL from the gradient leakage attack and MIA. To be specific, a simulated attacker (SA) network is embedded to defend the malicious attacks, and the gradient modification method is introduced to secure weight details. Moreover, the combination of FL and blockchain has been studied and proposed by many researchers to protect data privacy from attacks and poisoning \cite{article1111,li2022blockchain,wan2022privacy}. The work in \cite{liu2020secure}, a blockchain-based secure FL framework with smart contracts is proposed to defend against poisoning attacks. Additionally, a DP by adding well-designed Gaussian noises is introduced in smart contracts to prevent membership inference attacks. The fusion of FL and Blockchain technology is also used in UAV networks, the work in \cite{wang2020learning} proposed a secured FL-blockchain framework for UAV-assisted mobile crowdsensing. Local model updates are securely exchanged in blockchain-based architecture, and local DP is applied to protect the updated local models.



\subsection{Multi-agent Reinforcement Learning}
\subsubsection{Introductin of RL}
RL is one of the most essential branches of ML, which is mainly used to make decisions in a dynamic environment by interacting with the environment. The RL usually consists of 6 elements (agent, environment, policy, action, reward, and value function). In an agent-environment framework, the agent is the entity of the action, e.g., the UAV, and the agent performs the action and receives the reward through interaction with the environment. The process of RL can be seen as a game between the agent and the environment, which is also equivalent to a game between state and action. The policy is the mapping of each state to action. The value function represents how good a state is, and it is the total expected future reward from a given state. The aim of RL is to maximize cumulative rewards. Three popular algorithms Q learning, Deep Q Network (DQN) and Deep Deterministic Policy Gradient (DDPG) are going to be introduced. 

Q-learning \cite{watkins1992q} is a classic model-free RL, which based on Q-table to choose the action that can obtain the maximum reward as the next action. Q is Q (state, action), which is the expectation that taking action a can obtain reward in s state. The environment will feed back the corresponding reward r according to the action of the agent. Therefore, the main idea of the algorithm is to construct a Q-table of state and action to store and update the Q value. Then select the action with the highest Q-value as the next action.
\begin{figure}[t]
    \centering
    \includegraphics[width=0.5\textwidth]{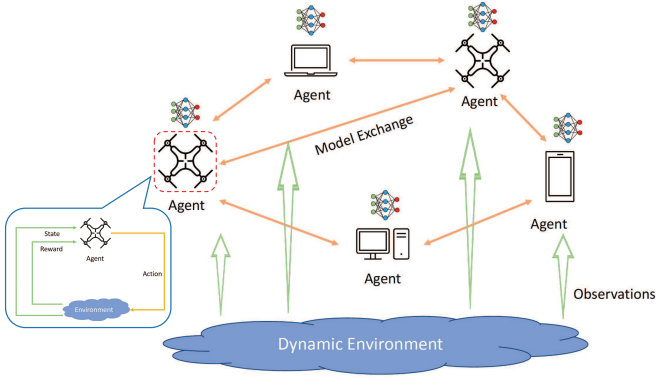}
    \caption{The framework of distributed RL \cite{letaief2021edge}.}
    \label{fig:RL}
\end{figure}

DQN is a deep RL algorithm that was proposed by Mnih et al. \cite{mnih2015human} in 2015. It is a value-based algorithm in which there is only the value functions network and no policy network, DQN is actually an improved version of Q learning, solving the problem of limited storage space for Q tables in Q learning. In DQN, a neural network is used to approximate the value function Q (state, action), using the state as input to the neural network, and Q (state, action) is computed by the neural network. The DQN then uses the $\varepsilon$-greedy strategy to output the action, $\varepsilon\in\mathbb[0,1]$ is a variable that trades off the exploration and exploitation. When the environment receives the action, it gives a reward and the next observation. Then it goes to update the parameters of the value function network according to reward. These processes are repeated until a good value function network is trained. Compared to Q-Learning, the DQN is improved by using a convolutional neural network to approximate the behavioral value function, whereas Q learning uses experience replay.The limitation of DQN is that it can only solve problems with discrete and low-dimensional action spaces. 

DDPG \cite{silver2014deterministic} is a model-free off-policy and actor-critic based algorithm, which has both a value functions network (critic network) and a policy network (actor network). In DDPG, the input to the critic network is action and observation, and the output is Q (state, action). The input of actor network is observation, and the output is action. When the environment gives an observation, the agent will make an action based on the actor network. After receiving this action, the environment will feed back a reward and an observation. It then updates the critic network according to the reward, and updates the actor network according to the recommendations of the critic network. The above process is repeated until a good actor network is trained \cite{lillicrap2015continuous}. DDPG helps the agent to find the optimal policy by maximizing the reward. The advantage of DDPG is that it can address high-dimensional continuous action space problems.

\subsubsection{Introductin of MARL}
MARL is often used in UAV swarm networks, because the single agent RL algorithm is a centralized algorithm where the central server implements to solve non-convex or time-dependent optimization problems \cite{9562559}. If used in UAV swarms, the increase in the number of UAVs will lead to a significant increase in the action space size and state space of single agent RL, which increases the information overhead, increases the complexity of training, and reduces the convergence speed. Therefore, it is necessary to propose distributed RL which can be executed by multiple devices together. The structure of distributed RL is shown in Fig.~\ref{fig:RL}.

The relationship of multiple agents in the MARL algorithm can be classified as fully cooperative, fully competitive, mixed cooperative and competitive, and self-interested \cite{canese2021multi}. Fully cooperative means that all agents are cooperative with each other which aims to maximize the common long-term returns. An example is self-driving cars, where they cooperate with each other to avoid collisions, ease traffic congestion, etc. In the full competitive setting, one agent's gain is the other's loss, and the sum of the total rewards received by all agents equals zero. Examples are chess and poker. In the mixed setting, agents can be both cooperative and competitive with other agents and present a sum reward. An example of this is in a football robotics game, where there is cooperation between the robots in the team and competition between the two teams. The self-interested setting means that each agent only considers maximizing its own interests, a typical example being automated stock trading systems.

MADRL is divided into three types according to how the learning models are trained and executed: fully decentralized, fully centralized, and centralized training with decentralized execution (CTDE)\cite{yun2021attention}. In a fully decentralized architecture, each agent implements a single RL independently and does not share any RL information between them, so each agent does not consider other devices when making decisions. Since no information is shared between agents, the convergence of MARL cannot be guaranteed. Furthermore, since each agent only considers its own optimal solution when making decisions, it is not able to find a locally optimal solution that maximizes the sum of the expected discounted reward of all agents. Fully centralized means that all agents send their observations to the central controller, which trains the model and makes decisions. MADDPG \cite{lowe2017multi} and G2ANet \cite{liu2020multi} belong to this framework. The advantage of this centralized framework is that the central controller knows all the actions, observations, and rewards, which can make better decisions to achieve the highest reward. However, the disadvantage is that it is slow during execution and cannot make real-time decisions, this is because communication is very time-consuming due to the frequent exchange of information about actions, rewards, status, etc. In CTDE MARL, the training phase is centralized by the central controller which knows all agent information, and the decision phase is made by each agent using its own policy network based on its own observations \cite{lowe2017multi}.

\subsection{Distributed Inference}
\begin{figure}[t]
    \centering
    \includegraphics[width=0.5\textwidth]{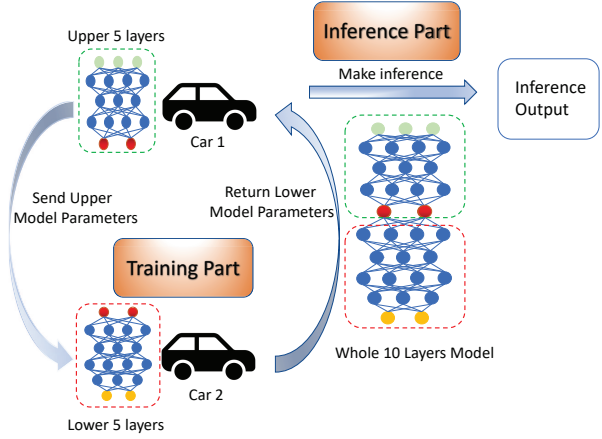}
    \caption{The framework of cooperative inference.}
    \label{fig:DI1}
\end{figure}

Inference is an important part of ML, where the first step is to train a model based on available data, and the second step is to apply new data on the model to make inference (regression or classification). Each device performs model training based on its own observations to obtain a local model, or local decision. Since each device observes a limited resource, which is one-sided and cannot make the final information, a server/cloud is needed to aggregate the model to make a final decision, this is called distributed inference. The distributed inference is divided into cooperative inference, model inference and decision inference. 

\subsubsection{Cooperative Inference} 
Cooperative inference involves dividing a large DNN into multiple parts, each  of which is executed by a helper, e.g., one part is trained on an end device and another part is computed by other edge devices or the cloud. For example, when a self-driving car 1 is performing collision avoidance prediction as shown in Fig.~\ref{fig:DI1}, car 1 trains a DNN model based on the data collected by its own camera. Due to its limited computing power and time constraints, car 1 needs the help of other cars around it to train the DNN model or upload it to the cloud if the model is large or there are no nearby cars. Suppose the DNN model has 10 layers, the first 5 layers are trained locally, and the last 5 layers are trained by the nearby car 2 or the cloud. During the training process, car 1 transmits the training result parameters of the first 5 layers to car 2, and car 2 returns the gradient information to car 1 after training the last 5 layers. Car 1 updates the model based on the gradient information and repeats the above operation until the loss function is small enough to complete the DNN model training. Finally, Car 1 makes an inference about whether there is an obstacle ahead. The work in \cite{teerapittayanon2017distributed} first proposed distributed DNN network models (DDNNs), where a DDNN can be jointly trained to do inference at a distributed computing level, which includes cloud, edge and end devices. In a DDNN, shallow layers of DNN can be performed quickly on the device and edges to do local inference. In other words, a shallow DNN is deployed on the end device, and then a large NN is deployed on the cloud for deep computation. At each inference query, multiple end devices first perform local inference through the shallow DNN based on their own collected piecewise data and then send the local inference results to the cloud, which performs deep processing and makes the final inference. This method enables adaptive decision making for offloading, which can meet the accuracy, communication and latency requirements of the target application. The DDNN reduces latency and communication costs (by more than 20 times) compared to traditional methods where the raw sensor data is offloaded to the cloud for processing. Inference accuracy is improved compared to fully local inference. In addition, privacy is protected as the edge device only sends the features of the intermediate computational model to the server.

\subsubsection{Decision Inference} 
\begin{figure}[t]
    \centering
    \includegraphics[width=0.5\textwidth]{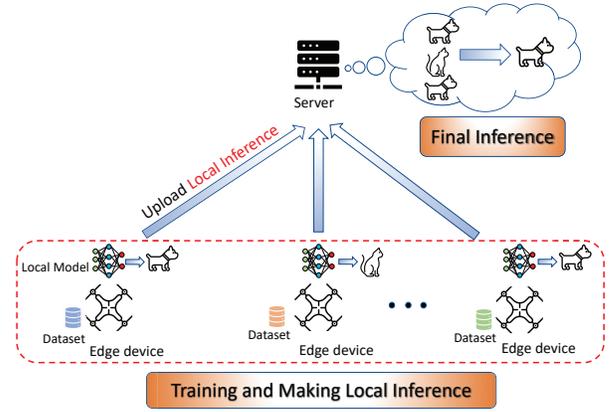}
    \caption{The framework of decision inference}
    \label{fig:DI 2}
\end{figure}

Decision inference refers to each edge device independently trains a DNN model based on the data it collects and then makes a local decision. Local decisions are then uploaded to the cloud and final inferences are made after aggregating the information. Since the data collected by each device is partial, it is not possible to make accurate inferences directly. For instance, applying a UAV swarm to identify objects as shown in Fig.~\ref{fig:DI 2},
the UAVs capture the same place from different orientations. Some UAVs may not capture the desired object at all, and some may only capture the corner of the object. Since each UAV captures images from different angles, the inference made by its locally trained model is very one-sided. It is necessary for the cloud to aggregate the local decisions of all UAVs to make the final classification.

\subsubsection{Model Inference} 
\begin{figure}[t]
    \centering
    \includegraphics[width=0.5\textwidth]{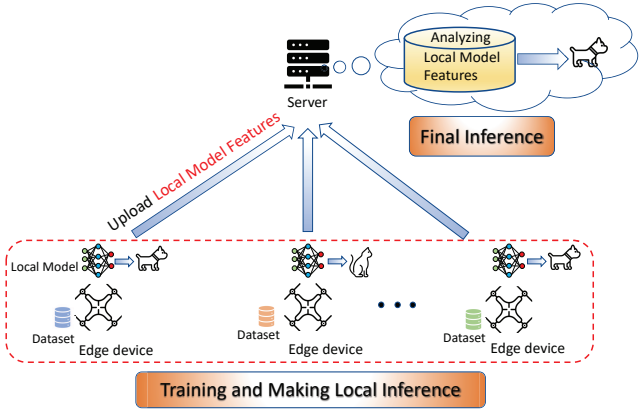}
    \caption{The framework of model inference}
    \label{fig:DI 3}
\end{figure}

The process of model inference is similar to that of decision inference as shown in Fig.~\ref{fig:DI 3}, 
except that after training the DNN model on each edge device, instead of uploading the decision directly to the cloud, the local model features are uploaded, and further detailed inference is made by the cloud.

In decision inference, only the local inference results are aggregated on the cloud, and the cloud cannot get too many detailed parameters to make decisions, so the error rate of inference will be high. On the contrary, in model inference, the model features are aggregated on the cloud, so the cloud can make detailed analysis and decisions.

\subsubsection{Advantages and Disadvantages of DI}
Distributed inference is more suitable for swarms of drones or networks of vehicles than traditional inference methods, since traditional inference method use cloud or edge servers to train model, where data needs to be transferred to the server, and both training and inference processes are executed by the server, resulting in high latency and communication burden. Moreover, sometimes UAVs may operate in harsh environments such as forests, deserts and oceans, where weather conditions may affect the quality of communication between the UAV and the remote server. Due to the height of the drone, high-rise buildings and other underground environments may also cause path loss, making it difficult to quickly and accurately upload the collected data to the cloud.
Distributed inference addresses these issues, allowing end devices to train models locally for inference, which greatly improves the communication burden and protects raw data privacy.

There is still an issue to be considered, as distributed inference on resource-limited edge devices can also be a challenge when DNN models are too large. For example, the ResNet-50 framework for image classification, which consists of 50 convolutional layers and about 100 megabytes, is difficult for edge devices with time constraints and limited memory and computational limitations. In particular, model inference and decision inference, both of which require full DNN model training to be done on the edge device. One promising approach is compression, adapting the size and complexity of the DNN to the edge device while minimizing accuracy loss. Pruning \cite{zhou2019edge} is one such approach, which reduces the complexity of the DNN and removes redundant weights that have little impact on performance. More advanced pruning algorithms are described in \cite{liu2020pruning}. Quantization \cite{han2015deep} is another method that aim to reduce the number of bits required to represent the network weights, rather than removing some weights. Sparsification as another effective method, by adding sparse constraints to the training process, it is straightforward to obtain a sparse network structure. Other studies have proposed binarization\cite{qin2020binary, courbariaux2016binarized, courbariaux2015binaryconnect}, where only binary numbers are used to represent the weights. In \cite{lin2015neural}, multiplication is converted to sign changes by randomly binarizing the weights, which does not compromise the classification performance and further simplifies the network operation. Therefore, it is possible to combine distributed inference and model compression when the DNN model is large.

\subsection{Spilt Learning}
Split learning (SL) is another distributed learning algorithm for training global learning network without sharing the raw data, which was proposed by Otkrist Gupta and Ramesh Raskar in 2018 \cite{gupta2018distributed}. SL involves splitting the ML model into several sub-models and training them distributed by multiple clients and a server. Each client trains the sub-models to the cut layer, and then sends the smashed data of the cut layer to the server. The server does the training for the remaining layers. Since in the cut layer, the client only transmits the output of the intermediate layer to the server in forward propagation and only the gradient is transmitted to the clients by the server in backward propagation, where no original data is involved, the privacy of the original data is protected.

\subsubsection{Introducation of SL}
\begin{figure*}[ht]
    \centering
    \subfigure[Horizontal partitioned data]{
    \label{fig:SL1}
    \includegraphics[width=0.5\textwidth]{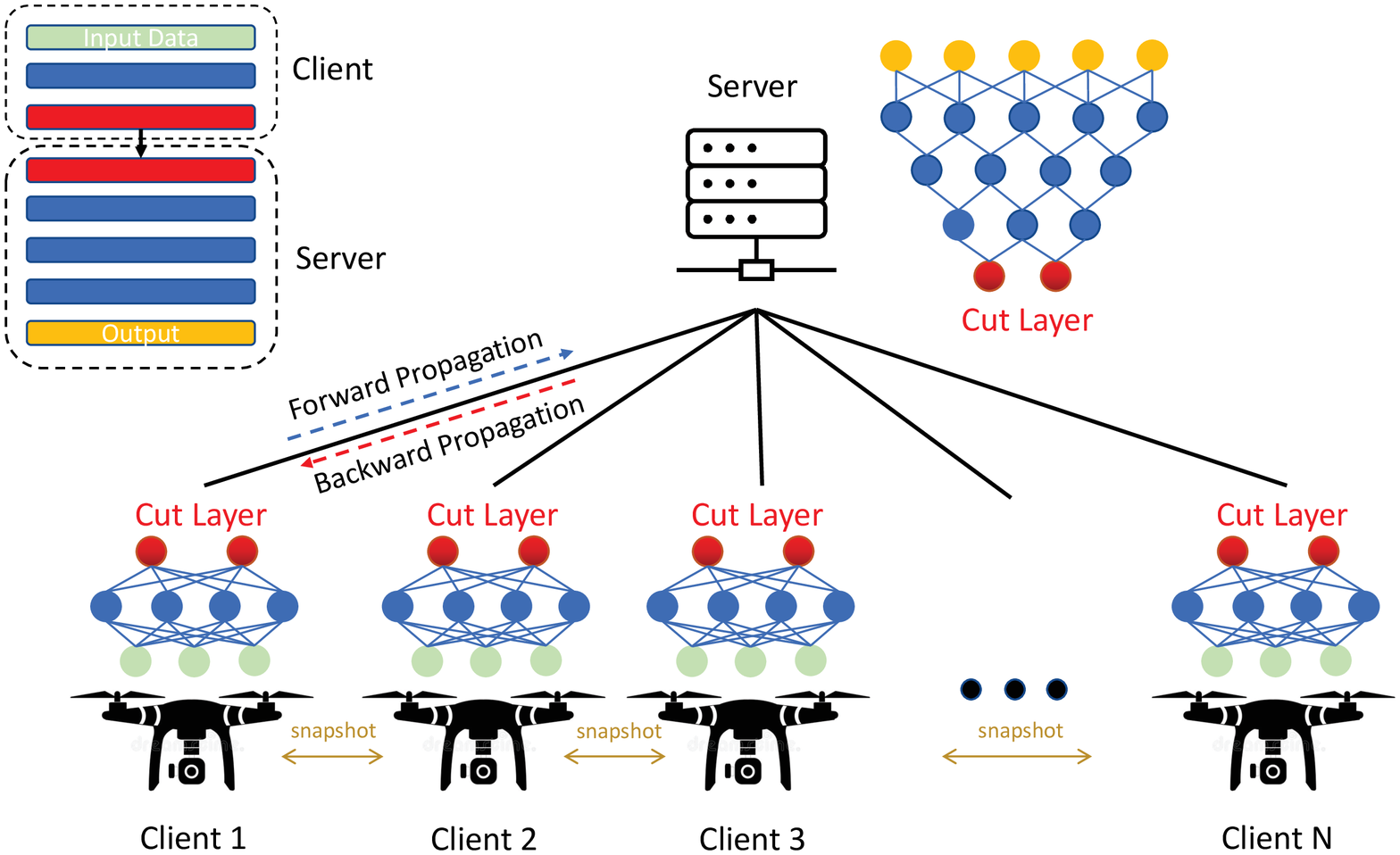}}\subfigure[Vertically partitioned data]{
    \label{fig:SL2}
    \includegraphics[width=0.5\textwidth]{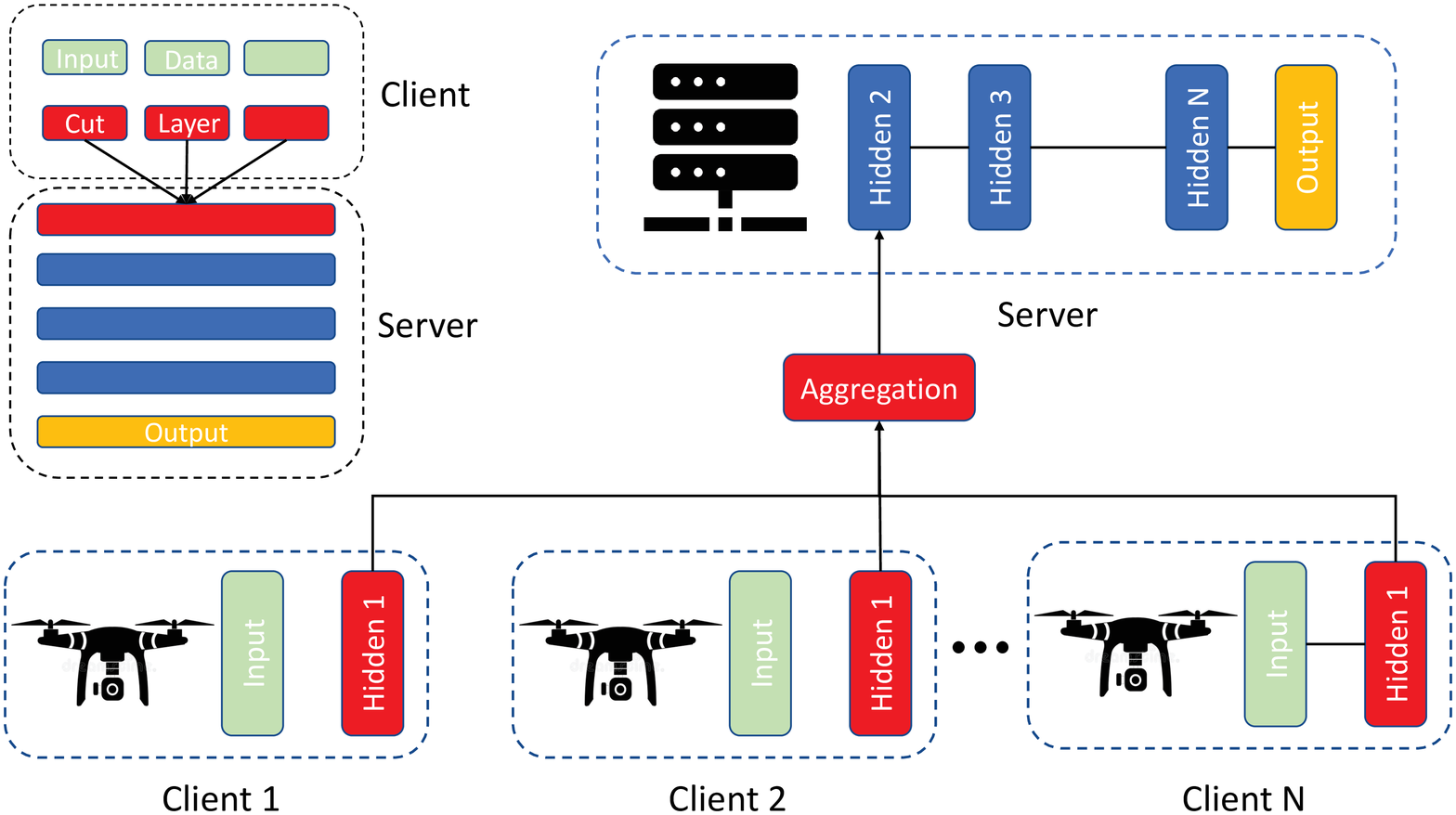}}
    \caption{Vertically partitioned data and horizontal partitioned data for split learning framework.}
    \label{1}
    \end{figure*}

SL models can be divided into horizontal data partitioning and vertical data partitioning \cite{vepakomma2018split}. Horizontal data partitioning as in Fig. \ref{fig:SL1} means that clients take turns alternating with the server. Specifically, each client trains a portion of the ML model up to the cut layer in sequence. e.g., there are $n$ clients in total, the first client trains a portion of the ML model to the cut layer with its own local data, and then the server completes the training of the rest of the layers, which completes the first round of forward propagation. The backward gradient propagation is done from the last layer on the server to the cut layer, and the gradient is sent to the client. The rest backward propagation is continued on the client-side and weights are updated. ML model is next trained by the second client, which downloads encrypted weights to the previous client before training, and repeats the above process until all clients and servers are trained. Vertical data partitioning as in Fig. \ref{fig:SL2} differs from horizontal book data partitioning in that it is not in a round-robin fashion, where all clients train the client-side model to the cut layer in parallel. The smashed data of the cut layer for all clients is then concatenated on the server-side as the input of the server and the rest sub-model is trained by the server. This process is repeated to complete forward and backward propagation until convergence.

SL has a wide range of applications which can be used in healthcare, communications, IoT and more. Vertically partitioned data for split learning is well suited for drones performing tasks on a large scale. For example, in \cite{ha2021spatio}, the authors propose the use of SL drones to monitor an entire city for any signs of fire. The impact of the number of drones and the imbalance of the data collected by the drones on SL was also investigated. The results show that the number of drones involved in SL does not affect the classification rate of the images and that a slight data imbalance has a higher resolution than a data balance, so for optimal performance, a slight data imbalance should be met.

\subsubsection{Advantages and Disadvantages of SL}
Compared to other distributed learning algorithms, such as FL, SL is splitting the NN into many parts, each part is trained by a different client and server, so each client does not need to train the whole NN \cite{krouka2021communication}, which greatly reduces the computational effort of the client and is very helpful for clients with limited energy and computational power, such as UAV. In \cite{singh2019detailed}, the authors compared the communication efficiency of SL and FL algorithms for small and large numbers of clients and for small, large and very large scale parameters. The experimental results illustrated that SL is more efficient and has higher scalability for large numbers of clients. In contrast, FL is more efficient under large-scale parameters, especially when the number of clients is small or the model size is small. The drawbacks of the SL algorithm are the increased computational load on the server and the challenge of determining the cut layer, where the computational power, data distribution and heterogeneous resources of different need to be taken into account in practice \cite{liu2022novel}. 

\section{Distributed Machine Learning Algorithms for Optimization Problems in UAV Swarms}\label{DS}

\begin{figure*}[htbp]
    \centering
    \includegraphics[width=1\textwidth]{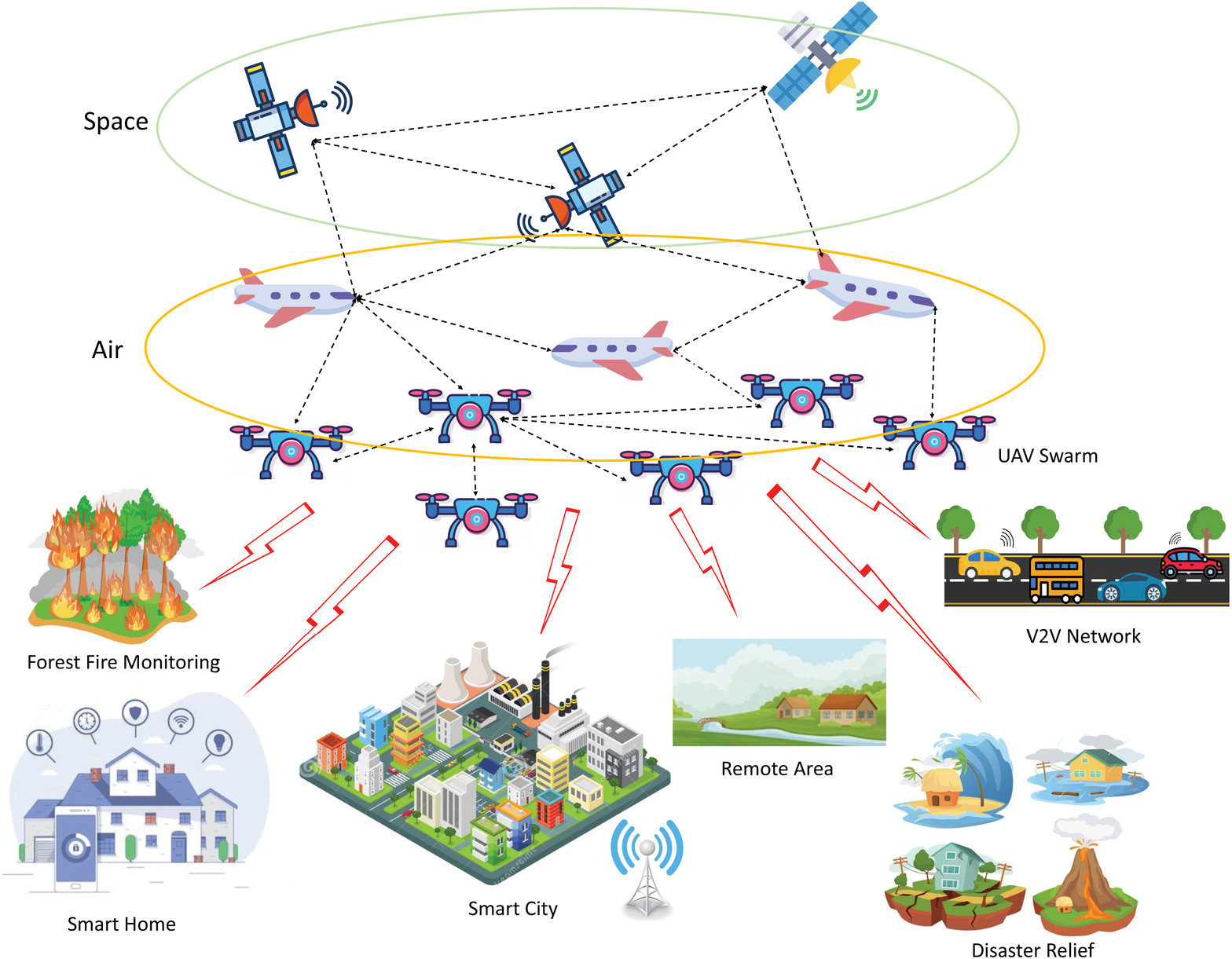}
    \caption{Some typical application scenarios of DL used in UAV swarms.}
    \label{fig:application}
\end{figure*}

In this section, based on recent research we shall summarize various optimization problems in UAV swarms, where optimizing UAV swarms by using DL or improving the performance of FL by optimizing the scheduling of UAV swarm. The optimization problems range from path optimization, wireless resource allocation, sensing, caching to computational resource allocation. In addition, non-ground network scenarios, such as satellite-UAV, are also included. The various application scenarios of DL used in UAV swarms are illustrated in Fig.~\ref{fig:application}.

\subsection{Trajectory Optimization}

Trajectory planning is an essential topic in UAV technology, which is a multi-constrained, mutually coupled multi-objective optimization decision problem \cite{9457160}. This main objective is to design low-cost flight paths that enable clusters of UAVs to accomplish their tasks collaboratively while avoiding collisions. In recent research, UAV path optimization in wireless networks has often been considered in conjunction with resource allocation, e.g., power control, target assignment, spectrum allocation, optimal interference, fairness, AoI, QoS, etc. Several recent studies with different scenarios are presented below, all of which use distributed learning for multi-UAVs trajectory optimization, with distributed reinforcement learning being more commonly applied.
In windy conditions, the offline path planning method may cause collisions or obstacles and it is not possible to plan an optimal path in time according to the current environment. Hamid et al. \cite{shiri2020communication} consider the real time path control for large amount UAVs in a windy environment. Due to a large amount of inter-UAV communication cost are needed when using common method. Mean field game (MFG) is applied to reduce the communication between UAVs. Solving the Fokker-Plank-Kolmogorov (FPK) and Hamilton-Jacobi-Bellman (HJB) equations is the main difficulty in real-time applications, and in order to overcome this challenge, they proposed a approximator based on NN, where every UAV runs two NNs, the outputs is the approximation solutions of FPK and HJB. To increase the NN training samples and accelerate the convergence of MFG, MfgFL-HF based on FL is proposed to enable share the model parameters between the UAVs, which not only takes into account some local non-observable samples, but also protects data privacy between UAVs.

In \cite{9195795}, authors proposed a novel time-constrained autonomous tracking framework based on multi-agent reinforcement learning for UAV swarms, which not only focuses on tracking accuracy but also considers searching time. Searching time is a crucial factor influencing learning performance, for instance, UAVs carry a limited battery, which limits their working time and prevents them from collecting data or interacting with the environment when it is out of power, thus affecting the training results. In order to minimize the searching time, the constrained Markov decision process is proposed as the flight decision model, which is solved by the multi-agent Q-learning (MQL) based tracking scheme. More specifically, in the proposed CA-MQL approach, Gaussian Process (GP) is applied to estimate the reference point location. Compared with other strander Q-learning and MQL, the proposed MQL shows a better performance in tracking rate and searching time. 

The work in \cite{venturini2021distributed} consider applying UAV swarms in the area of monitoring and exploring. As UAV swarms are applied to different scenarios, the models of the scenarios are different, and the algorithms need to be relearned, which leads to a huge waste of computational resources. Authors proposed a flexible trajectory control scheme based on Multi-Agent Reinforcement Learning (MARL) which can be easily switched from one scene to another without much additional training. Moreover, this framework can handle the non-uniform distribution of targets and obstacles. In proposed scheme, they model the environment as a Networked Distributed Partially Observable Markov Decision Process (ND-POMDP) and adopt Distributed Deep Q-Learning (DDQL) algorithm. The experimental results show that the algorithm has good transfer capability, compatibility, and detection speed.

Qin et al. \cite{qin2021distributed} proposed a distributed trajectory control framework based on multi-agent reinforcement learning for multiple UAV-BSs named MAUC, which considers fairness communication in user-level. This algorithm adopts centralized training and distributed execution. Simulation results demonstrate that MAUC can enhance the fairness of communication by sacrificing a small amount of throughput under certain circumstances. However, because MAUC requires some prior knowledge to train the model, it can only be used in specific scenarios and is not flexible. 

Klaine et al. \cite{klaine2018distributed} researched the fast-positing multiple UAVs as mobile base stations to supply reliable communication service in emergency situations, such as earthquake, typhoon or other large-scale disasters. They proposed a RL method based on distributed Q-learning to determine the optimal position of UAVs, which main objective is to rapidly deploy UAVs and maximize network coverage. Numerical results illustrate that this proposed algorithm can find best position in dynamic environment and outperforms than other fixed approaches.    

Khamidehi et al. \cite{8904880} also investigated the trajectory planning of multiple aerial base stations (ABSs) in communication network, which aim is to find the optimal ABSs trajectory that maximizes the number of users covered by each ABSs. To achieve above goal, they realize that in addition to optimal trajectory design, subchannel allocation and optimal power are also two essential factors should be considered, which can support user obtain higher data rates. Thus, they divide this complex problem into two subproblems: trajectory optimization, and joint subchannel and power allocation. Then, they proposed a distributed Q-learning method to tackle above subproblems. Simulation results show that although Q-learning is a model-free reinforcement learning approach, it is able to obtain optimal trajectory of ABSs by receiving reward signal with information from the network topology.

The work in \cite{8846699} investigated a joint collaborative optimization problem in target assignment and path planning of multi-UAV (MUTAPP) in a certain scenario, where a UAV swarm was needed to fly to area which has targets distributed in different locations and exists some fixed threat area that are inaccessible to the UAVs. In addition, the UAVs are required to have the shortest total flight distance and avoid collisions between them. The above problem can be seen as a combinational optimization with target assignment and path planning. Qie et al. proposed a simultaneous target assignment and path planning (STAPP) to deal with the above optimization problem in dynamic environment, which is based on multi-agent deep deterministic strategy gradient (MADDPG) algorithm and is a type of MARL. In SATAPP, MUTAPP problem is considered as a Markov decision process (MDP) and constructed as MA system. The MADDPG was then used to train the system to solve both goal assignment and path planning according to the corresponding reward structure. Experimental results demonstrate that the STAPP can effectively settle the MUTAPP in dynamic environment as it only requires the locations of the UAVs, target, and threat areas for execution. 

Liu et al. \cite{8727504} designed a novel framework for the trajectory design and power control of multiple UAV to improve user throughput. They proposed a three-step method. Firstly, proposing a MAQL based placement algorithm to determine the initial optimal position of UAVs. Secondly, using eco state network (ESN) based prediction algorithm to predict the future position of mobile users. Finally, MAQL based approach was applied to determine the trajectory acquisition and power control for the UAVs. In \cite{challita2018deep}, Challita also proposed a deep reinforcement learning algorithm based on ESN to optimal interference-aware path planning for multi-UAVs.

Hu et al. \cite{hu2018reinforcement} investigated the trajectory design of multi-UAVs which perform real-time sensing tasks in a cellular network. Firstly, a sense-and send protocol was proposed to coordinate the multi-UAVs, and nest Markov chains are applied to estimate the performance of this protocol. Then, they proposed the enhanced multi-UAV Q-learning approach to tackle the UAV trajectory design problem. Similarly, Hu et al. \cite{hu2020cooperative} also considered the optimal trajectory design for the cellular Internet of UAVs, where multiple sensing tasks were executed by UAVs continuously through cooperative sensing and transmission, with the aim of minimizing the age of information (AoI) of the tasks accumulated during a period. They proposed a distributed sense-and-send protocol to coordinate the multi-UAVs. Based on this proposed protocol, the trajectory design problem was formulated as an MDP, and a compounded-action actor-critic (CA2C) algorithm based on the deep reinforcement learning (DQN and DDPG) algorithms to solve the MDP, which is capable of handling agents whose actions involve both continuous and discrete variables.

In \cite{9188401}, authors investigated path planning for the complex scenarios of UAVs mixed operation and proposed a UAV collaborative path planning approach based on MAXQ multi-agent hierarchical RL to improve the environmental adaptability and self-coordination of UAVs under the complex situation of mixed operation. 

The work in \cite{wu2020cellular} considered trajectory design problem of a cellular Internet of UAVs with overlaying UAV-to-Device (U2D) communications to guarantee the Quality-of-Service (QoS) for sensing services. Trajectory design problem was regarded as a Markov decision problem (MDP) and a DQN-based multi-UAV DRL algorithm was proposed to solve this problem. Simulation results point out that the proposed algorithm outperform than single-agent algorithm and policy gradient algorithm in terms of total utility.

 In \cite{wang2020multi}, authors proposed a multi-UAVs assisted mobile edge computing (MEC) framework, where each UAV with different trajectories is controlled by an agent and support the user equipment (UEs) on the ground. In order to jointly maximise the geographical fairness among all the UEs, the fairness of the UE-load of each UAV and at the same time minimise the total energy consumption of the UAV, a multi-agent based DRL trajectory control algorithm called multi-agent deep deterministic policy gradient (MADDPG) is proposed.

 A decentralized MARL method is considered to solve the multi-target tracking guidance (MTTG) in unknown environments of decentralized UAV swarms in \cite{wenhong2021improving}. They propose a maximum reciprocal reward based approach to make UAVs learn cooperative tracking policies in a decentralized manner. In particular, the reciprocal reward refers to reshaping the original reward of UAVs, which is the dot product of the reward vectors of all neighboring drones and the corresponding dependency vector between the drone and its neighbors. The Pointwise Mutual Information (PMI) neural network is utilized to obtain the relationship between UAVs directly. Moreover, they proposed the experience sharing Reciprocal Reward Multi-Agent Actor-Critic (MAAC-R) algorithm to learn the collaborative sharing policy for UAV swarms. Simulations indicate that the proposed MAAC-R algorithm outperforms than benchmark approach in improving cooperation and inspires a variety of cooperative UAV swarm tracking behaviors.
 
Ant colony optimization (ACO) is a bio-inspired swarm optimization algorithm, which is one of the mostly used path planning algorithms. However, ACO has some drawbacks about premature convergence and stagnation problems. To overcome above problems, Cekmez et al. \cite{cekmez2016multi} proposed multi-colony ACO approach for obstacle avoidance UAV path planning, where multiple ant colonies seek an optimal solution by exchanging valuable information with each other.

Based on recent literature related to DL for trajectory optimization for UAV swarms, one can easily observe that distributed RL is an interesting technique to solve the real-time path planning of UAV, such as distributed Q-learning. Table \ref{tab:trajectory} summarizes all works in this survey related to the trajectory optimization of UAV swarms using DL. Although constraints and optimizations vary from scenario to scenario, the goal of most path planning is to complete data collection tasks or provide services to ground users in an efficient and energy-saving manner while avoiding collisions.

\begin{table*}[htbp]
\centering
\caption{DL applications for trajectory optimization in UAV swarms}
\label{tab:trajectory}
\begin{tabular}{|c|c|c|c|}
\hline
Reference&  DL Methods& Application&    Objective \\
\hline
\cite{shiri2020communication}&   FL+NN &\begin{tabular}{p{4.5cm}} Real time path control in a windy environment \end{tabular} & \begin{tabular}{p{4.5cm}} Avoid collisions or obstacles\end{tabular} \\
\hline
\cite{9195795}&   MARL (Q-learning)&\begin{tabular}{p{4.5cm}}	Time constrained autonomous tracking \end{tabular}&\begin{tabular}{p{4.5cm}}	Reduce searching time\end{tabular} \\
\hline 
\cite{venturini2021distributed} &   MARL(DDQL)&\begin{tabular}{p{4.5cm}} Flexible trajectory control for exploration and surveillance\end{tabular}&\begin{tabular}{p{4.5cm}}	Find the targets without colliding with each other or with obstacles
\end{tabular} \\
\hline
\cite{qin2021distributed}&   MARL(DRL)&	\begin{tabular}{p{4.5cm}}Trajectory control\end{tabular}&\begin{tabular}{p{4.5cm}}	Enhance the fairness communication in user-level\end{tabular}  \\
\hline
\cite{klaine2018distributed}&   Distributed Q-learning& \begin{tabular}{p{4.5cm}}	Fast positing in emergency situations\end{tabular}& \begin{tabular}{p{4.5cm}}	Rapidly deploy UAVs and maximize network coverage \end{tabular}\\
\hline
\cite{8904880}&   Distributed Q-learning& \begin{tabular}{p{4.5cm}}Trajectory planning of ABSs\end{tabular}&\begin{tabular}{p{4.5cm}}	Maximize the number of users covered by each ABSs \end{tabular}\\
\hline
\cite{8846699}&   MADDPG&\begin{tabular}{p{4.5cm}}	Target assignment and path planning\end{tabular}&\begin{tabular}{p{4.5cm}}	Shortest total flight distance and avoid collisions\end{tabular} \\
\hline
\cite{8727504}&   MAQL based ESN&\begin{tabular}{p{4.5cm}}	Trajectory design and power control\end{tabular}&\begin{tabular}{p{4.5cm}}	Improve user throughput \end{tabular}	\\
\hline
\cite{challita2018deep}&   Distributed DRL based ESN&\begin{tabular}{p{4.5cm}}	Path planning\end{tabular}&\begin{tabular}{p{4.5cm}}	Optimal interference-aware\end{tabular} \\
\hline
\cite{hu2018reinforcement}&   MAQL&	\begin{tabular}{p{4.5cm}}Trajectory design for real-time sensing tasks\end{tabular}&\begin{tabular}{p{4.5cm}}	Propose a sense-and send protocol\end{tabular}	\\
\hline
\cite{hu2020cooperative}&   Multi-agent deep RL&\begin{tabular}{p{4.5cm}}	Optimal trajectory design for the cellular Internet of UAVs\end{tabular}&\begin{tabular}{p{4.5cm}}	Minimize the AoI of the tasks accumulated\end{tabular} \\
\hline
\cite{9188401}&   Multi-agent hierarchical RL&\begin{tabular}{p{4.5cm}}	Path planning for complex scenarios\end{tabular}&\begin{tabular}{p{4.5cm}}	Improve the environmental adaptability and self-coordination \end{tabular}\\
\hline
\cite{wu2020cellular}&   MARL(DQN)&\begin{tabular}{p{4.5cm}}	Trajectory design for cellular Internet of UAVs\end{tabular}&\begin{tabular}{p{4.5cm}}	Guarantee the QoS for sensing services \end{tabular}\\
\hline
\cite{wang2020multi}&   MADDPG&\begin{tabular}{p{4.5cm}}	Trajectory control in UAV assisted MEC\end{tabular}& \begin{tabular}{p{4.5cm}}	Jointly maximize the geographical fairness among all the UEs\end{tabular}	\\
\hline
\cite{wenhong2021improving} & MARL &\begin{tabular}{p{4.5cm}} MTTG in unknown environments of decentralized UAV swarms  \end{tabular} &\begin{tabular}{p{4.5cm}} Improve the cooperation of homogeneous UAV swarms \end{tabular}\\
\hline

\cite{cekmez2016multi}&   ACO&\begin{tabular}{p{4.5cm}}	Path planning\end{tabular}&\begin{tabular}{p{4.5cm}}	Avoid obstacle  \end{tabular}\\
\hline
\end{tabular}
\label{tab1}
\end{table*}

\subsection{Wireless Resource Allocation}
In addition to the trajectory planning of the UAV swarm, wireless resource allocation of UAVs is another essential part, which includes delay, beamforming, frequency, coding and power control. In this part, many types of resource allocation optimization problems are solved by distributed learning in UAV swarm scenarios. The common resource allocation problem is NP-hard, as multiple conflicting conditions need to be considered simultaneously, such as low overhead, low latency, high throughput, etc \cite{bithas2019survey}.

\subsubsection{Beamforming}

Xu et al. \cite{xu2020transmit} considered a beamforming design and optimization for multi-UAVs, where UAVs with limited power can be too far away from BSs and other UAVs to communicate directly when they are on mission. An effective way to extend the communication range of the UAVs is to transmit beamforming. Thus, they proposed a transmit beamforming algorithm based on PSO for multi-UAVs (MUTBF) to make the transmit beam narrower by adjusting the heading angles of the UAVs. The simulation results illustrate that the proposed MUTBP approach outperform than single MSNR and adaptive JRSB, narrowing the bandwidth while ensuring a high SINR.

The work in \cite{xiao2020fully} proposed a fully decentralized FL framework with an inexact stochastic parallel random walk alternating direction method of multipliers (ISPW-ADMM) to solve any full decentralized FL tasks over time-varying graphs, while maintaining high learning efficiency and improving privacy. To further validate the fast convergence and high communication efficiency of the proposed framework, they present a concrete on-board task in which the robust beamforming (BF) design of an extreme learning machine (ELM)-based FL model for UAV swarm communication is investigated.

\subsubsection{Resource Block Allocation}

The work in \cite{shamsoshoara2019distributed} considered the task allocation problem for spectrum shortage in UAV networks. They studied a relay-based cooperative spectrum leasing scenario where multi-UAVs work together on a remote sensing mission. In this scenario, UAVs are divided into relaying UAVs and sensing UAVs. The relay UAVs provide relaying services for a ground-based primary user in the network to obtain spectrum access for the sensing UAVs which work is to transmit their collected data to a fusion center. The task allocation algorithm based on distributed MARL (Q-learning) was proposed to assign UAVs tasks based on maximizing the total utility of the system, whether as relaying UAVs or as sensing UAVs. The simulation results demonstrated that the proposed algorithm was verified its convergence in different scenarios.

In \cite{cui2019multi}, authors investigated the real-time resource allocation of multi-UAVs enabled communication network to maximize the long-term reward, which can be formulated as a stochastic game. Each UAV can be seen as a learning agent and the objective of them is to find a strategy of the resource allocation to maximizing its expected reward. Therefore, they proposed a MARL based dynamic resource allocation approach to tackle the above stochastic game, where each agent execute a decision algorithm independently based on Q-learning. The simulation results point out that this proposed algorithm can obtain a good tradeoff between system performance and information exchange overheads.

Zhang et al. \cite{zhang2018matching} considered the resource allocation and distributed power control in the dense non-orthogonal multiple
access based UAVs network, and proposed a two-stage optimization scheme to mitigate interference and data traffic congestion. To be specific, the centralized resource allocation problem was formulated as a roommate matching problem, which was solved by time slot allocation approach. The distributed power allocation was formulated as a mean field game (MFG) and solved by finite difference method, which optimizes the network spectrum efficiency. These two processes reduce the impact of interference between users and optimize the spectral efficiency of the network respectively. Simulation results show that this proposed algorithm can significantly improve the communication reliability of dense UAV networks.

In \cite{9500304}, Azmy et al. considered the optimal distribution of data in a heterogeneous UAVs network with different computing and communication capabilities, which can affect the training time of FL models and thus the performance of the tasks performed. They proposed an FL-UAV scheme to handle the above problems. Specifically, they first determine the optimal distribution of data among UAVs to minimize the FL learning time, and then they use discrete optimal transport over the UAV trajectory to find the point that minimizes D2D communication time for exchanging data to reach the optimal distribution of data.

The work in \cite{yin2021resource} investigated the problem of resource allocation and trajectory design for a swarm of UAVs providing base station services to ground users with unavailable addresses and channel parameters via frequency division multiple access. They propose a MARL-based approach to address resource allocation and trajectory design with the aim of optimizing the overall throughput and fair throughput. In particular, the parameterized deep Q-network (P-DQN) is used as a local critic network for each UAV, and QMIX is applied to aggregate local critics, and an entropy-like fairness indicator is applied as a reward in RL. Moreover, a distributed framework is proposed to optimize the overall throughput, where each drone can provide its gradient to train the global model, and the training process can be performed in parallel. Simulation results show that the proposed MARL algorithm can automatically plan the path without knowing the user location and channel parameters. Furthermore, almost the similar throughput is achieved for ground users.

In \cite{9428021}, Jouhari et al. considered how to enable a UAV with limited computing and memory resources to perform one-site CNN inference while performing a mission in a low-latency application, instead of sending it to the server for inference. They proposed a distributed CNN inference method to allocate CNN on UAVs. Specifically, UAV swarms collaboratively execute a CNN model, dividing a request into different UAVs for computation, each UAV calculating one layer, and then sending the results of the intermediate layer to the next UAV. The ultimate optimization goal is to minimize total lantency to execute the decision-making. In addition, they also consider the impact of UAV mobility on air-to-air communication latency and introduce the UAV mobility prediction method.

In \cite{9838867}, a novel distributed learning structure, hybrid split and federated learning (HSFL), is proposed to alleviate the communication overhead and protect user data privacy in wireless UAV networks with limited computation capacity and different data distributions over UAVs. To be specific, in HSFL architecture, a portion of UAVs with large dataset train entire ML model locally via FL algorithm, and another portion of UAVs with weak computation capacity train the ML model via SL algorithm in collaboration with BS. Compared with the state-of-the-art DL algorithms, FL and SL, the simulation results showed that the HSFL algorithm provides higher learning accuracy than FL and reduces communication overhead than SL under non-IID data.
\subsubsection{User Allocation}

Authors in \cite{nguyen2019real} considered a framework for real-time deploying UAVs as relays to quickly recover the communication network in case of disaster. They proposed a K-means based user clustering selection model for UAV networks and a distributed resource allocation algorithm with low computational complexity and fast convergence to maximize the end-to-end sum rate. These proposed algorithms help a large number of users rapidly recover communication connections in a disaster situation.

To improve the downlink wireless coverage of UAV swarms during search and rescue missions in the unknown area, the authors considered a distributed strategy in \cite{9762762}, where each UAV maximizes wireless coverage by exchanging only local information, and the wireless coverage problem is divided into several distributed optimization subproblems. They proposed a UAV swarm wireless coverage game on an undirected random graph to solve the above non-convex subproblem. In addition, they propose a DL algorithm to obtain the optimal Nash equilibrium. The simulation results show that the proposed algorithm improves the coverage by 58 \%.

In \cite{yang2019application}, the task allocation of UAV clusters was considered. To achieve cooperative work between UAVs in a UAV cluster, Yang et al. proposed a distributed RL based task allocation for UAV clusters. In this algorithm, the UAV can automatically and dynamically adjust the mission strategy via calculating the task performance efficiency. To be specific, they first proposed a networking scheme and using expansion strategy to settle the problem of initial state of UAV networking. Then, DRL was applied to deal with the dynamical allocation problems of wireless channel, thereby optimising the delay of UAV data transmission. Moreover, an instance was given of how the above approaches can be applied to solve the task scheduling in UAV cluster.

\subsubsection{Power Control}

In FL training, proper client scheduling and resource allocation improve the learning performance, but the battery energy of the client UAV will affect the working time of the UAV (it is impractical for the running UAVs to replace the battery), which further influents learning performance. The concept of simultaneous wireless information and power transfer (SWIPT) was proposed in \cite{8214104}, which means that information and energy can be transmitted simultaneously to wireless devices through radio frequency (RF) signals.The work in \cite{9613446} proposed the framework of joint client scheduling and wireless resource allocation of FL in a micro-UAV swarm enabled by SWIPT. In the UAV network, BS as the server and multiple UAVs as clients. UAVs use their collected data sets to train  local models and transmit them to BS, and then BS aggregates the received model to update the global model. Finally, both the ML model and power are transmitted to UAVs from BS. UAVs rely on battery power and harvested energy to repeat the above operations. Moreover, they improve the performance of FL by maximizing the percentage of scheduling UAVs, which can be addressed by transforming to a convex mixed integer nonlinear programming (MINLP) problem. 

It is difficult to maintain a continuous connection between the UAV swarms and the base station using a centralized ML approach, and when the amount of data is large, the traffic load in the UAV network also increases. So, Zeng et al. \cite{zeng2020federated} applied FL algorithm in UAV swarm, which is the first work to apply FL for UAV swarms. To be specific, a UAV swarm consists of one leading UAV and some following UAVs, where leading UAV as central server and following UAVs as clients. They cooperatively train a FL model.  Each client UAVs trains a local model by using its own collected data, and then sends local model to server. Server will aggregate received local model and update a global FL model which will send back to clients. They also researched the influence of wireless factors and other uncertain factors on the convergence of FL algorithms for UAV swarm, such as transmission delay, fading, wind and mechanical vibrations. To accelerate FL convergence with low energy consumption and stability of the swarm’s control system, they proposed a joint power allocation and scheduling design to optimize the convergence performance of FL. Simulation results demonstrate that this algorithm performs well on the convergence of FL, saving 35\% communication rounds compared with baseline.

In \cite{shen2022joint}, authors investigated the joint optimization of training and resource allocation in UAV swarms to minimize energy consumption. They proposed FL based optimization algorithm with low complexity to solve the minimization of overall training energy consumption of UAV swarm and the minimization of maximum energy consumption of UAV swarm. In this optimization approach, they considered the fairness of energy consumption among different UAVs, which can be formulated as a min-max optimization problem and can be solved by proposed approach with adding an auxiliary variable. The simulation results demonstrated that the proposed algorithm outperforms than other four baseline schemes in reducing the energy consumption of the UAV swarm. Moreover, the work in \cite{9264742}investigated the energy efficient computation and transmission resource allocation for FL in wireless communication networks and proposed an iterative algorithm to solve it. UAVs are not considered in this work, but this optimization method can be applied in UAV swarms.

In \cite{liu2019price}, authors proposed a price-based downlink power allocation framework for multi-UAV wireless networks and modeled the interaction between UAVs and ground users as a Stackelberg game. In Stackelberg game, UAVs as the leaders, maximize their own revenue by choosing the optimal power price, and each ground user maximizes its own utility by choosing a power strategy. However, there is a problem of equilibrium program with equilibrium constraints (EPEC) formed by Stackelberg game. They investigated the lower equilibrium of the ground user and proposed distributed iterative algorithm to solve it.

Table \ref{tab:multicol} summarizes recent works related to the wireless resource allocation of UAV swarms. The biggest limit to the deployment of UAVs in various applications is limited power. A proper resource allocation mechanism can make energy-constrained UAVs accomplish tasks efficiently. 


\begin{table*}[htbp]
\caption{DL applications for wireless resource allocation in UAV swarms}
\begin{center}
\begin{tabular}{|c|c|c|c|c|}
 \hline
    Category & Ref. & DL Methods & Application &  Objective \\
    \hline 
    \multirow{2}{*}{\begin{tabular}{p{2cm}}Beamforming\end{tabular} }&\cite{xu2020transmit} & PSO &\begin{tabular}{p{4.5cm}} Beamforming optimization for multiple UAVs \end{tabular} &\begin{tabular}{p{4.5cm}} Narrow the bandwidth of transmit beam to extend the communication range \end{tabular}\\
    \cline{2-5}
    & \cite{xiao2020fully} & FL &\begin{tabular}{p{4.5cm}} Robust beamforming design for UAV swarm communication \end{tabular}&\begin{tabular}{p{4.5cm}} Solve any full decentralized FL tasks over time-varying graphs, while maintaining high learning efficiency and improving privacy\end{tabular}\\ 
    \hline 
    \multirow{5}{*}{\begin{tabular}{p{2cm}}Resource Block Allocation\end{tabular}}&\cite{shamsoshoara2019distributed} & MARL(Q-learning) & \begin{tabular}{p{4.5cm}}Spectrum management in UAV networks\end{tabular} &\begin{tabular}{p{4.5cm}} Maximize the total utility of the primary user and the UAV network\end{tabular}\\ 
    \cline{2-5}
     & \cite{cui2019multi}  & MARL(Q-learning)  &\begin{tabular}{p{4.5cm}} Real-time resource allocation\end{tabular} & \begin{tabular}{p{4.5cm}}Maximize long-term rewards\end{tabular}\\
     \cline{2-5}
     &  \cite{zhang2018matching} & Distributed MFG  &\begin{tabular}{p{4.5cm}} Time and frequency resources allocation in the NOMA-based UAVs\end{tabular} &\begin{tabular}{p{4.5cm}} Reduce system interference and improve system reliability\end{tabular} \\
    \cline{2-5}
     &  \cite{9500304} & FL  &\begin{tabular}{p{4.5cm}} Resource allocation in a heterogeneous UAVs swarm with different computation and communication capabilities \end{tabular} &\begin{tabular}{p{4.5cm}} Minimize the learning time of FL\end{tabular} \\
      \cline{2-5}
     &  \cite{yin2021resource} & MARL(P-DQN,QMIX)  &\begin{tabular}{p{4.5cm}} Resource allocation and trajectory design in aerial base stations based UAVs\end{tabular} &\begin{tabular}{p{4.5cm}} Optimize the overall throughput and fair throughput\end{tabular} \\
      \cline{2-5}
     &  \cite{9428021} & Distributed inference  &\begin{tabular}{p{4.5cm}} Distributed CNN inference in resource-constrained UAV swarms  \end{tabular} &\begin{tabular}{p{4.5cm}} Minimize the decision-making lantency\end{tabular} \\
      \cline{2-5}
     &  \cite{9838867} & FL and Split learning  &\begin{tabular}{p{4.5cm}} Resource allocation for aerial user equipments   \end{tabular} &\begin{tabular}{p{4.5cm}} Alleviate the communication overhead and protect user data privacy\end{tabular} \\
    \hline
     \multirow{2}{*}{\begin{tabular}{p{2cm}}User Allocation\end{tabular}}&\cite{9762762}  & Distributed learning &\begin{tabular}{p{4.5cm}} Wireless coverage optimization for UAV swarms in search and rescue missions   \end{tabular} & \begin{tabular}{p{4.5cm}} Maximize the downlink wireless coverage \end{tabular} \\
    \cline{2-5}
    & \cite{yang2019application} & Distributed RL &\begin{tabular}{p{4.5cm}}  Real-time task allocation within a UAV cluster in dynamic environment \end{tabular} &\begin{tabular}{p{4.5cm}}  Enable UAV to automatically and dynamically adjust the mission strategy\end{tabular} \\
    \hline 
     \multirow{3}{*}{\begin{tabular}{p{2cm}}Power Control\end{tabular}} & \cite{9613446} & FL  &\begin{tabular}{p{4.5cm}} Joint client scheduling and wireless resource allocation of FL in an SWIPT enabled micro-UAV swarm \end{tabular} &\begin{tabular}{p{4.5cm}} Maximize the percentage of scheduling UAVs to improve the performance of FL \end{tabular}\\
     \cline{2-5}
     & \cite{zeng2020federated} & FL  &\begin{tabular}{p{4.5cm}}Joint power allocation and scheduling design of UAV swarms\end{tabular} &\begin{tabular}{p{4.5cm}} Minimize the number of communication rounds required for convergence\end{tabular}\\
    \cline{2-5}
     & \cite{shen2022joint} & FL  & \begin{tabular}{p{4.5cm}}Joint training and resource allocation optimization in UAV swarms\end{tabular} & \begin{tabular}{p{4.5cm}}Minimize the energy consumption\end{tabular}\\
    \hline

\end{tabular}
\end{center}
\label{tab:multicol}
\end{table*}

\subsection{Sensing, Caching and Computing Resource Allocation}
In this part, we focus on DL algorithms used in sensing, caching, and computing resource allocation of UAV swarms. UAV swarms collect data from the environment by sensing, some popular data can be cached local, and others can be sent to relays for storage, which is convenient for users to download. When the information in the data is large, the UAV can process it itself, or send it to an edge server or cloud for processing if it is unable by itself.

\subsubsection{Sensing and Monitoring}

Liu et al. \cite{liu2020federated} proposed a monitoring and forecasting air quality sensing framework for UAV swarms based on FL. In the airborne sensing system, haze photos are taken by UAV sensing and the lightweight Dense-Mobile model is used to predict the AQI inference. In the ground sensing system, a graph convolutional NN based long short-term memory (GC-LSTM) is proposed to predict AQI inference in real time. Both learning processes use FL method. Simulation results show that this framework outperform than existing approaches and this work is one of pioneering examples of air quality forecast by using FL.

Zhang et al.\cite{9210077} considers using multi-UAV system to classify images in exploration scenarios, especially in places that cannot be easily reached, such as mountain tops and seas. FL-based multi-UAV system can fuse data collected by different UAVs and train the global CNN model to classify images, which can also reduce the communication cost between UAVs and servers. Ground fusion centre (GFC) is used here as a server to coordinate UAVs. In the proposed FL-assisted image classification method, each UAV first trains a local model based on its own collected data, then sends it to GFC for aggregation into a global CNN model, and finally global model is sent back to each UAV for the next round of local model updates. In order to reduce the computational complexity of GFC and speed up the global update at the GFC, they propose a weighted zero-forcing (WZF) transmit precoding (TPC) scheme. Simulation results show that the proposed method guarantees high classification accuracy at low communication cost. 

In \cite{lee2022federated}, the authors considered using UAV swarms to perform gas sensing on mountains to find the origns of gas resource and consider building swarm intelligence to enable UAV swarm to be more autonomous. The federated reinforcement learning (FRL)-based UAV swarm aerial remote sensing system is proposed, which uses proximal policy optimization (ppo) as the RL algorithm to ensure high UAV autonomy and FL to train the model to ensure the privacy of local data. With the combination of FL and RL, it makes swarm intelligence more reliable and robust. Simulation results demonstrate that the proposed system performs better in terms of learning performance and is more suitable for UAV swarms from various perspectives than existing RL-based centralized systems.

To address the large overhead caused by transmitting large amounts of raw data during centralized ML model training in UAV swarm networks, Cui et al. \cite{9562134} proposed a new distributed online learning approach where each UAV can conduct ML model training locally with its own data. In particular, a UAV is selected at the beginning of each iteration to receive the latest model parameters, then the model is trained using local data, and the model parameters are transmitted to the UAV selected in the next iteration.
The selection of the UAV will affect the effectiveness of model training, so they proposed a metric, Age of Updates (AoU), that quantifies the freshness of data and the contribution in the learning process to better select UAVs for training.

\subsubsection{Caching}

In order to solve the problem of increased delay in backhaul links due to congestion and high activity in 5G networks\cite{lahmeri2021artificial}, caching at small base stations has become a promising method. Caching the popular information among users on the base station allows for fast interaction of the required content with the user, which not only improves the throughput of users but also decreases transmission latency \cite{7539325}\cite{8269405}. However, fixed base stations cannot efficiently provide services for mobile users. When mobile users move to a new cell, the new base station does not have the cache of their requested content. In this case, flexible base stations, such as cache-enabled UAVs, are a promising new solution for tracking mobile users and delivering their requested content efficiently.

In order to maximize the throughput in the device-to-device (D2D) caching network, the authors in \cite{9473012} investigated the trajectories of multiple cache-enabled UAVs that dynamically serve ground users with unpredictable mobility pattern. They propose a 3D trajectory design based on the cooperative MARL algorithm to achieve the above purpose. Specifically, each UAV determines its optimal 3D trajectory by cooperative multi-agent Q-learning in a distributed manner. Simulation results show that the proposed method is more effective in improving network throughput than traditional single and multi-agent Q-learning algorithms.

In \cite{9060924}, authors investigated the content caching approach for 6G integrated aerial-terrestrial network which include heterogeneous base stations (hgNBs) consisted of UAVs and terrestrial remote radio heads. In order to mitigate the privacy of the UEs, they proposed a distributed heterogeneous computing platform (HCP) and 2-stage FL algorithm. The local training model of the content experience and preferences of the UEs and hgNBs is first built, and then the model is uploaded to the HCP controller, which trains its global model using an asynchronous parameter update algorithm based on the proposed 2-stage FL with CNN. After that, the global model is returned to the UEs and hgNBs for the next round of local model updates. Overall, 
the collaborative communication capabilities of UEs and hgNBs are leveraged by HCP-based FL platform to jointly predict content caching placement via considering localized content popularity, traffic distribution and UE mobility.


\subsubsection{Computing}

Dhuheir et al. \cite{ye2020dynamic} investigated a novel decentralized DQN (dDQN) algorithm for dynamic coordination in UAV swarm assisted mobile edge computing (MEC) where multiple UAVs are applied to help terrestrial edge servers supply better edge computing service \cite{8764580}. In this system, each UAV can observe the current state, and a task-heavy UAV can offload a part of its buffer of computational tasks to an adequately selected neighbor. The key challenge of the above process is how to coordinate communication and computation among these UAVs dynamically in a dynamic environment, which can be solved by proposed dDQN with low complexity. The simulation results showed that the proposed algorithm can help UAV swarms learn good dynamic coordination strategies so as to achieve noticeably higher mission computation rates.

The works in \cite{chen2021joint} considered relative delay optimization in MEC-assisted coalition-based UAV swarm networks, where UAVs are divided into coalitions depending on the requirements of the task, and  a leader and several members are consisted in each coalition. They jointly optimize computational offloading, channel access and scheduling to minimize total heterogeneous data delay in MEC-assisted UAV networks. The impact of scheduling approaches on delay was analyzed and a shortest effective job first (SEJF) based scheduling approach was proposed to shorten the relative delay. Furthermore, the joint optimization problem in distributed UAV swarm is formulated as a game model and is proved to be an exact potential game (EPG) that admits at least one pure strategy Nash Equilibrium (PNE). Therefore, they propose a concurrent best-better response (CBBR) based offloading algorithm to reach the proposed PNE. The simulation results illustrated that the CBBR has fast convergence speed. In addition, the proposed SEJF based scheduling method reduces the delay by up to 30 percent compared to other scheduling methods. Compared with the works in \cite{chen2021joint}, deployment is considered in \cite{yao2020distributed}. Yao et al. investigated the joint deployment, power control, computation offloading and channel access optimization problem in MEC-assisted coalition-based UAV swarm. To be specific, they consider the selection of UAVs as mobile MEC servers in the UAV swarms consortium to alter channel quality and thus improve network performance. A UAV swarm is divided into a number of federations, each of which has a head as a server, and the rest as members to collect data. They propose a distributed algorithm, RAN-BBR, to jointly optimize the deployment of the federated UAV swarm and to compute offloading, power control and channel access problems. The special feature of the algorithm is that multiple policies are compared in each iteration, thus not only avoiding the need to traverse all policies but also speeding up convergence. The simulation results illustrated that distributed RAN-BBR has greater energy savings than the approaches without deployment optimization. The average energy savings were 31\% under different time constraints and over 20\% under different coalition amounts.

In \cite{9449289}, the authors investigated a distributed collaborative computing framework in UAV swarms to achieve real-time image processing in dynamic environments. They first proposed a computing collaboration structure to collaboratively compute CNNs, where the coordinator schedules the computational resources of other UAVs and the UAVs are formed into different groups to process different images in a pipelined manner. The objective of this approach is to minimize batch execution time, which is modeled as a nonlinear programming model that jointly optimizes the computational load and communication consumption among the UAVs. To address this NP-hard issue, the distributed CNN based resource scheduling algorithm based on DQN is proposed to improve the real-time low latency processing capability. Simulation results show that the proposed method significantly outperforms other schemes in terms of reducing the execution time and improving the utilization of UAV computational resources. Compared with the unified scheduling scheme and single UAV scheme, the batch execution time is approximately 20\% and 40\% less, respectively.

Table \ref{tab:computing} summarizes the contributions in this survey related to sensing, caching and computing of UAV swarms. 
According to this table, FL is widely used for both sensing and caching related tasks. As a special type of FL, FRL can be used in aerial remote sensing. To realize real-time application, the distributed online learning algorithm can help establish robust scheme for UAV swarms. For caching task, the caching placement related resource allocation problem can also be solved using MARL. For computing task, distributed CNN and CBBR algorithms are good candidates for joint computing and communication problem.  

\begin{table*}[htbp]
\caption{DL applications for sensing, caching and computing resource allocation in UAV swarms}
\begin{center}
\begin{tabular}{|c|c|c|c|c|}
    \hline
     Category & Ref. & DL Methods & Application &  Objective \\
    \hline
    \multirow{4}{*}{Sensing}&\cite{liu2020federated} &FL & \begin{tabular}{p{5cm}} Monitoring and forecasting the air quality in a fine-grained manner    \end{tabular}&\begin{tabular}{p{4.5cm}} Achieve high accuracy, low energy consumption and privacy-preserving AQI predictions\end{tabular} \\
    \cline{2-5}
     & \cite{9210077} & FL  & \begin{tabular}{p{5cm}} Image classification in exploration scenarios\end{tabular} &\begin{tabular}{p{4.5cm}} Minimize the cost of communication between the UAV and the GFC and the computational complexity of the GFC\end{tabular}\\
     \cline{2-5}
     &\cite{lee2022federated} & FRL &\begin{tabular}{p{5cm}} FRL-based UAV swarms system for aerial remote sensing \end{tabular} &\begin{tabular}{p{4.5cm}} Establish robust and reliable SI for UAV swarms\end{tabular}\\
     \cline{2-5}  
     &\cite{9562134} & Distributed online learning &\begin{tabular}{p{5cm}} UAV swarms as sensors to collect data and train ML model in the edge computing framework \end{tabular} &\begin{tabular}{p{4.5cm}} Establish robust distributed online learning scheme for UAV swarms\end{tabular}\\
    \hline
    \multirow{2}*{Caching} & \cite{9473012} & MARL(Q-learning) &\begin{tabular}{p{5cm}} Cache-enabled UAVs as flying BS to serve ground users in aerial-terrestrial wireless caching network \end{tabular} &\begin{tabular}{p{4.5cm}} Maximize  the throughput in the D2D caching network\end{tabular}\\
    \cline{2-5}
     & \cite{9060924} & FL &\begin{tabular}{p{5cm}} Collaborative cache optimisation among the UEs, UAVs and HCP\end{tabular} &\begin{tabular}{p{4.5cm}} Predict the content caching placement and guarantee the QoS\end{tabular} \\
    \hline 
    \multirow{4}*{Computing}&\cite{ye2020dynamic} & dDQN  &\begin{tabular}{p{5cm}}Dynamic coordination in MEC-assisted UAV swarms\end{tabular}&\begin{tabular}{p{4.5cm}} Reduce the complexity of finding the best dynamic coordination strategy\end{tabular} \\
    \cline{2-5}
     & \cite{chen2021joint} & CBBR &\begin{tabular}{p{5cm}}Joint computing offloading and channel access optimization issues based on SEJF scheduling for MEC-assisted UAV networks\end{tabular} &\begin{tabular}{p{4.5cm}} Minimize total heterogeneous data delay\end{tabular}\\
     \cline{2-5}
     & \cite{yao2020distributed}& RAN-BBR &\begin{tabular}{p{5cm}}Joint deployment, power control, computational offloading and channel access optimization issues for MEC-assisted UAV swarms\end{tabular}  &\begin{tabular}{p{4.5cm}} Maximize convergence speed and minimize computational cost\end{tabular}\\
    \cline{2-5}
     & \cite{9449289} & Distributed CNN+DQN  &\begin{tabular}{p{5cm}} Computing resource allocation of UAV swarms for real-time sensing tasks \end{tabular} &\begin{tabular}{p{4.5cm}} Minimize batch execution time \end{tabular}\\
    \hline
    
\end{tabular}
\end{center}
\label{tab:computing}
\end{table*}

\subsection{Massive Aerial Access for UAV Swarms}
The previous studies have been related to terrestrial communications, where UAVs can act as aerial base stations to provide internet services to users where the internet is not available on wired infrastructure. With the development of 6G cellular networks, terrestrial networks are still the main component of cellular networks. Due to the high cost of building terrestrial infrastructure such as base stations, network providers often focus on densely populated areas, while remote rural and mountainous areas still do not have adequate broadband coverage. In addition, the reliability and responsiveness of existing cellular network infrastructure is not high in some cases, such as when faced with a natural disaster, connectivity outages will lead to the inability to communicate with the outside world in a timely manner, resulting in property damage and even loss of lives \cite{9275613}. 

One promising solution to increase the resilience of networks is to dense cellular sites \cite{9275613}, but this will face many problems such as increased operational costs and the complexity of the terrain when the network is deployed. To address these problems and promote high-capacity network services worldwide, some companies are attempting to use non-terrestrial networks (NTNs) to provide network services, which is a part of 6G current focused research. 

In \cite{yoo2021federated}, the authors considered using satellites and the UAV framework to provide communication resources to users in densely populated areas. Satellites cover a wide range of areas, but their orbit are fixed, and their paths cannot be changed randomly. As a result, satellite resources are limited, and it is difficult to cope with dynamic environmental changes such as user movement and Internet use over time. To solve the above problems, the authors propose an adaptable aerial access network (ANN) system which consists of low-Earth orbit (LEO) satellites and Federated Reinforcement Learning (FRL) based UAVs to increase the capacity of network service. In proposed system, UAVs can automatically detect where communication resources are scarce according to the network traffic map and find the optimal location according to the dynamic traffic distribution on the ground via training with the FRL algorithm. Simulation results show that the system can provide network services in different areas, even in areas where the demand for network resources changes rapidly. Moreover, using the system in densely populated areas, the UAV helps the satellite to share some network traffic load, so as to reduce the load on the satellite and enable the satellite to provide more resources for rural areas. Compared to the satellite-only ANN, the proposed AAN provides 3.25 times more communication resources and 5.1 percent lower latency than a satellite-only AAN.
\section{New Applications for UAV Swarms}\label{New}
In this section, we shall introduce some state of art application for UAV swarms, such as RIS-enabled UAV swarm, and discuss the challenges of DML used in these applications.

\subsection{RIS-enabled UAV Swarm Networks}
With the development of 6th generation networks, there is a need to develop more spectrum and energy-saving and low-cost technologies \cite{9497709}. Reconfigurable intelligent surface (RIS) has been introduced as a new generation technology to enhance energy efficiency and provide a huge amount of spectrum for network \cite{basar2019wireless}. RIS is a planar surface that enables to optimize signal reflection by adjusting the phase shifts and amplitudes of passive reflecting elements on the surface to achieve fine-grained reflection-beamforming \cite{shang2021uav}. Some recent studies \cite{9367542}\cite{9777886} have considered installing RIS on UAVs to achieve three-dimensional signal reflection, which provides higher deployment flexibility, reliable air-to-ground links, and 360° panoramic full-angle reflections \cite{lu2020enabling} than terrestrial RIS. In \cite{zhang2019reflections}, the authors propose an RL-based approach for efficient deployment of UAV-IR (a UAV-carried intelligent reflector) to reflect downlink mmWave to mobile users without line-of-sight (LoS) channels. However, due to the limited battery capacity and payload of UAV, a large number of reflective elements cannot be carried on a single UAV, resulting in the scalability of the large aperture gain of the aerial RIS (ARIS) supported by a single UAV cannot be guaranteed \cite{9777886}.

To solve the above dilemma, UAV swarms can be used in combination with RIS to further improve the performance of ARIS \cite{9579401}.Fig.~\ref{fig:RIS} illustrates the application of SARIS consisting of UAV swarms-enabled ARIS in a wireless network. The specific scenarios are as follows. SARIS can provide 360° reflection for ground users when they are in a blind spot, and UAV swarms-RIS can provide multiple data streams when there are too many users or a large amount of data. SARIS can also be deployed to provide ubiquitous wireless connectivity and reliable data transmission. For example, in areas not covered by cellular networks such as tunnels and underpasses. SARIS can increase signal coverage to provide wireless connectivity and reliable data transmission to users. It can also provide over-the-air computing services for users. Moreover, SARIS can enhance the signal capacity to provide stable wireless communication for dense users. In V2V networks, SARIS reflects interference signals back to eavesdroppers, providing continuous secure communications.

For the beamforming design of SARIS, in \cite{liu2020machine}\cite{yang2020deep}, authors used the DRL algorithm to accomplish the decision problem through interacting with the dynamic environment by trial and error. However, the DRL approach is not well-suited for highly dynamic systems as it requires some time to converge. For SARIS trajectory design, the UAV swarm trajectory is challenging to derive mathematically as it is a three-dimensional problem and is related to a time sequence. Nevertheless, it is possible to use distributed RL to learn a suboptimal solution, similar to the use of distributed RL to plan UAV swarm trajectories in Chapter 3, except that the application scenarios and constraints are different and beam formation strategies need to be considered here. In MARL, each UAV as an agent to learn its 3D trajectory by offline trial and error, while the RL-based trajectory design should also consider the phase shift of the reflecting elements on each RIS, as this facilitates the UAV to select an action for passive beamforming from a limited action space. Furthermore, the work in \cite{yang2020federated} developed a RIS-FL framework for smart IoT (including UAV swarms) based on communication efficiency of over the air computing (AirComp), where two new techniques RIS and AirComp are proposed to address the limited communication bandwidth challenge in the aggregation phase of FL models. In particular, RIS enhances signal strength by reconfiguring the wireless propagation environment to reduce model aggregation errors.

\begin{figure*}[!h]
  \centering
    \subfigure[Multiple data stream]{\includegraphics[width=0.49\textwidth]{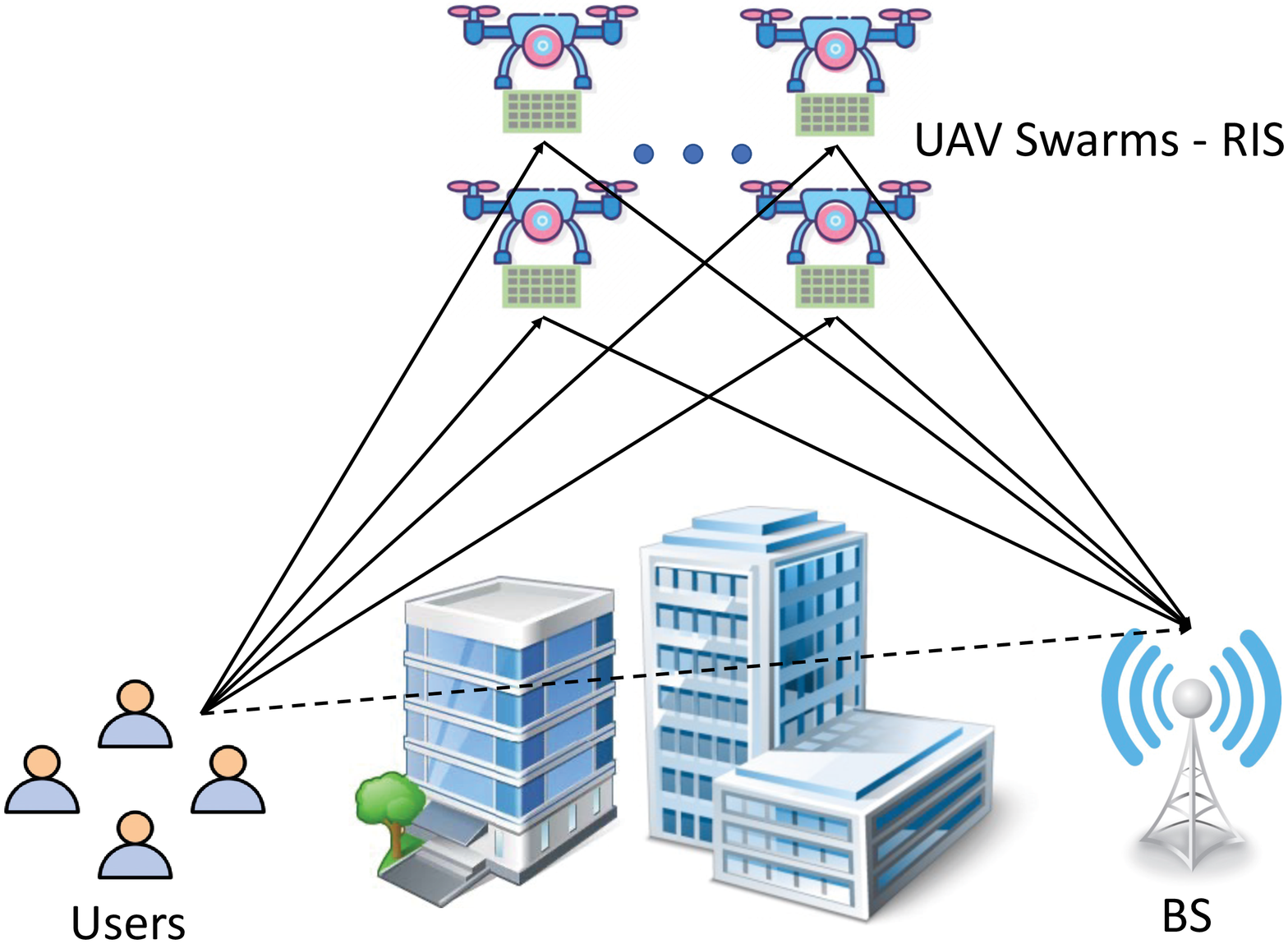}} 
    \subfigure[Enhance signals coverage]{\includegraphics[width=0.49\textwidth]{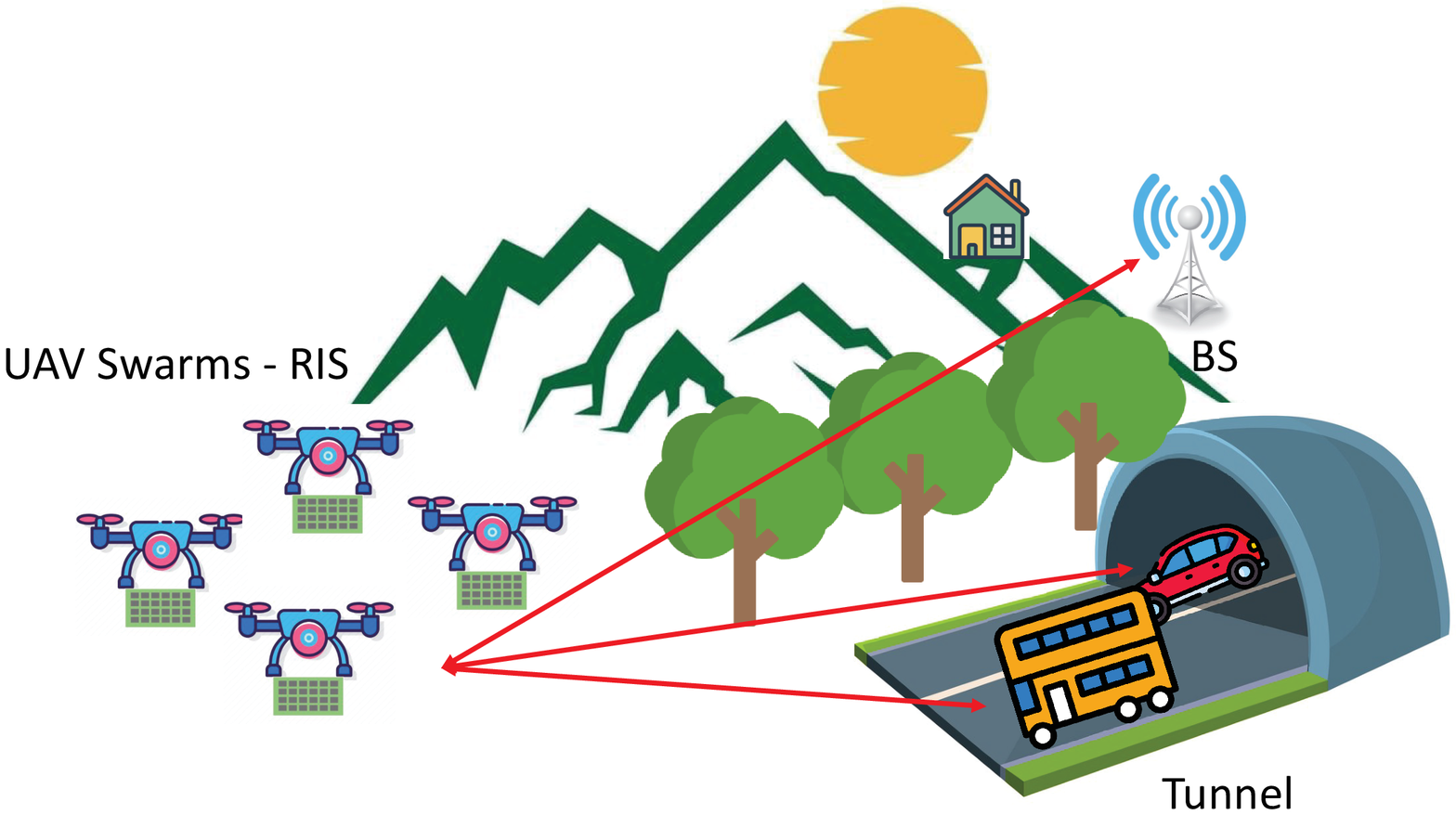}} 
	\subfigure[Over the air computing]{\includegraphics[width=0.49\textwidth]{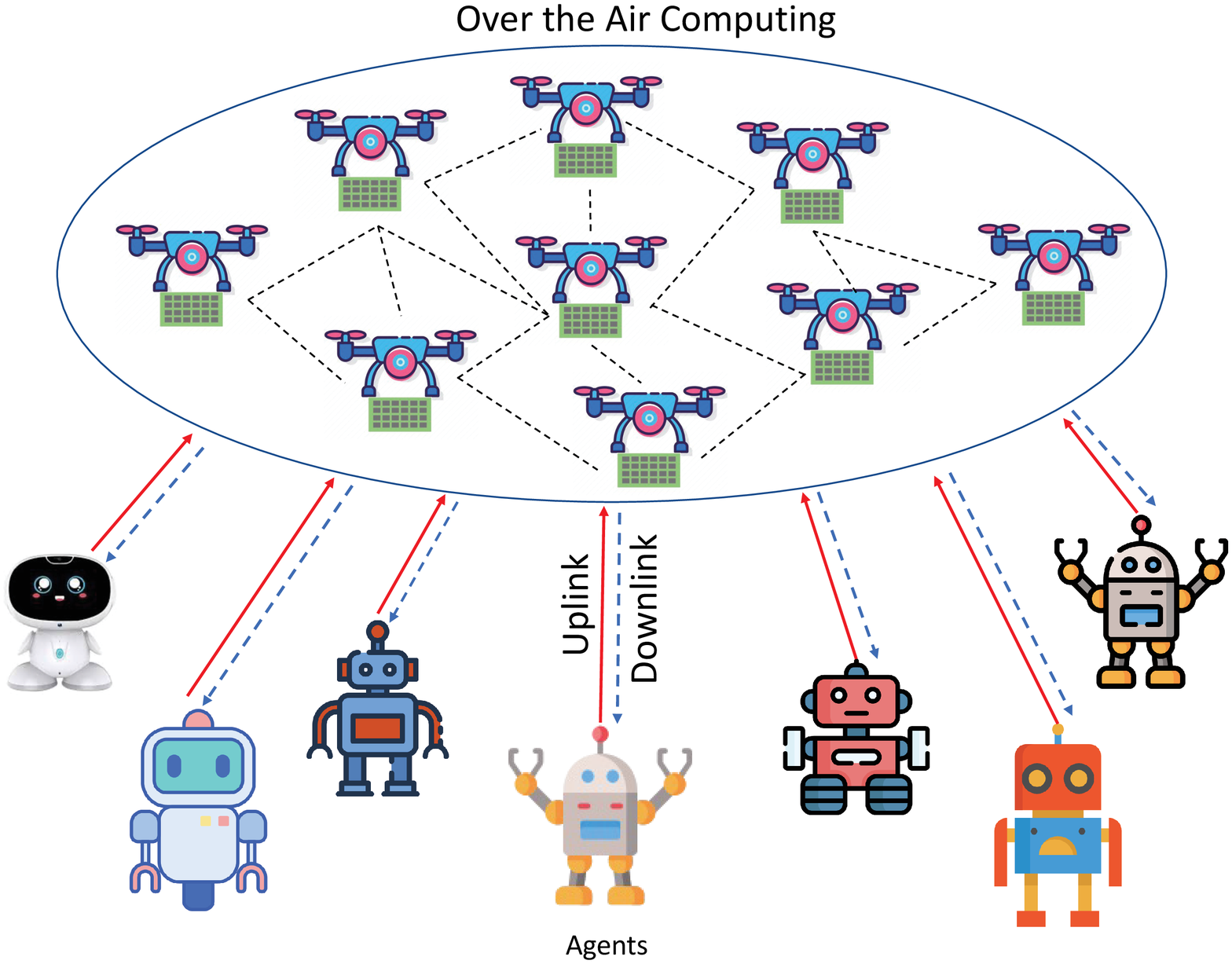}} 
    \subfigure[Increase signal capacity]{\includegraphics[width=0.49\textwidth]{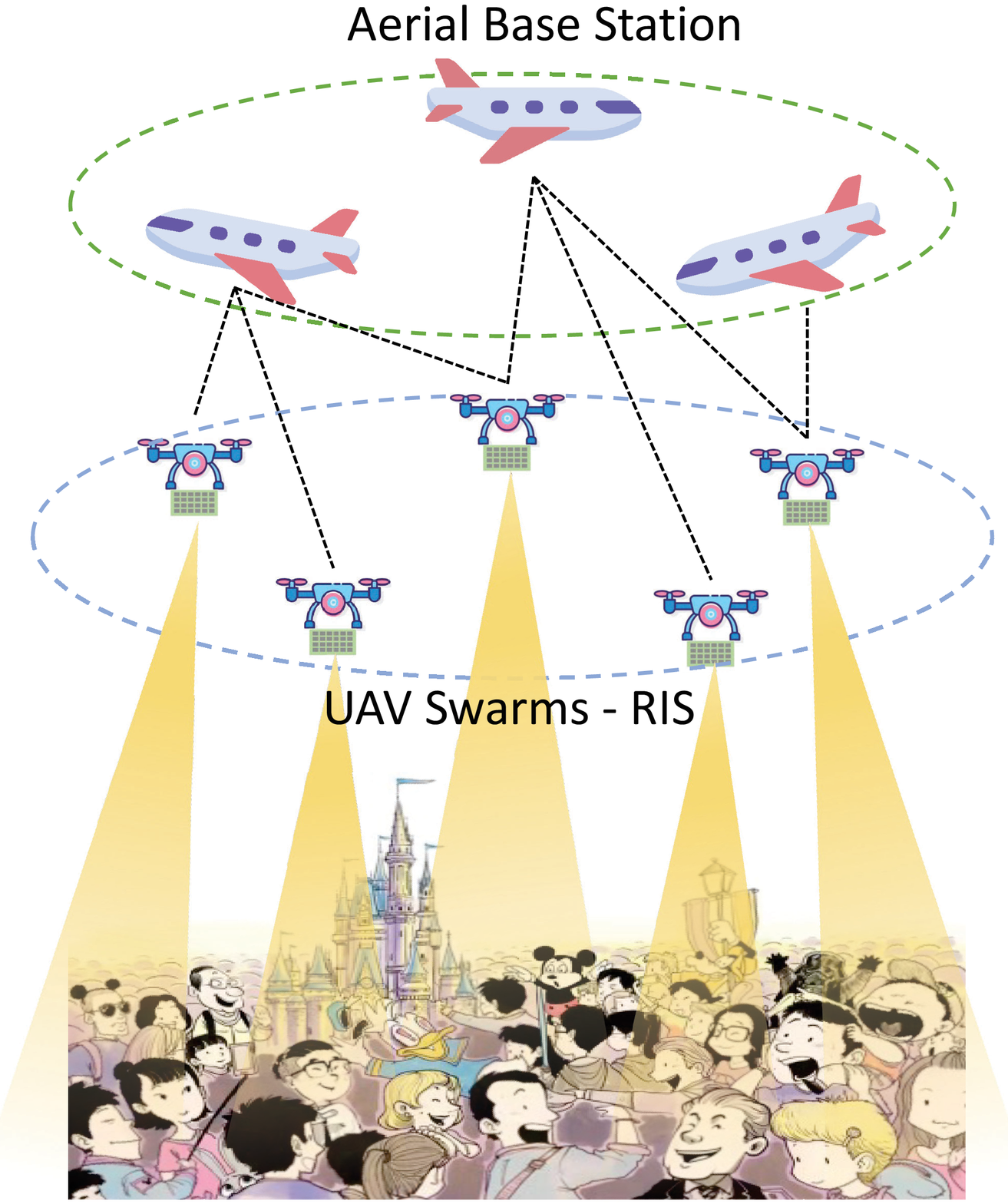}} 
	\subfigure[Trackable physical layer security]{\includegraphics[width=0.5\textwidth]{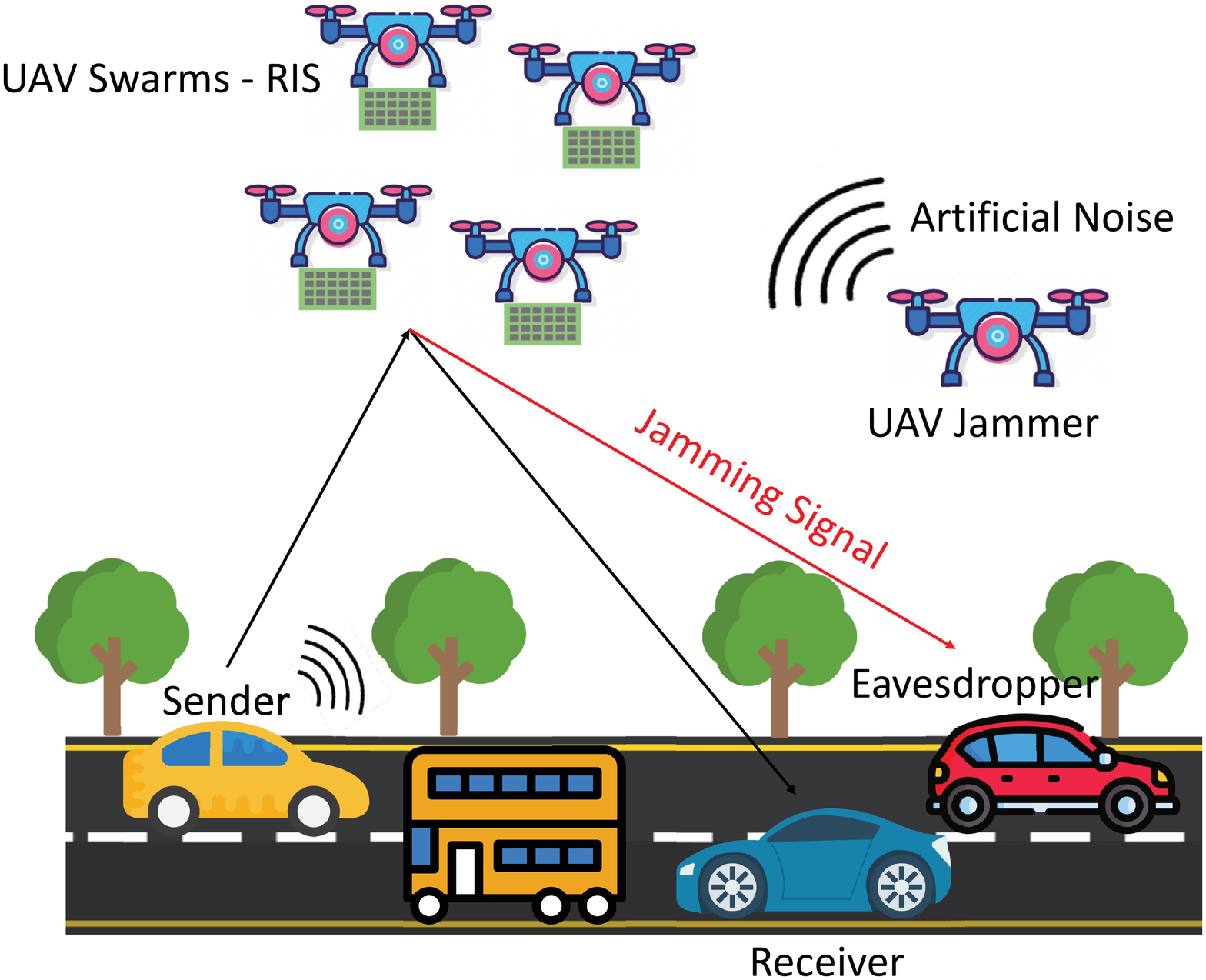}}

  \caption{Applications of UAV swarm-enabled RIS}
	\label{fig:RIS}
\end{figure*}

\subsection{Semantic-UAV Swarms}
As communication and AI technologies continue to converge, the Internet of Everything has been seen as one of the key 6G visions, where semantic communication and Edge Intelligence are expected to be two key drivers \cite{letaief2021edge}. Semantic communication is a new form of communication based primarily on AI and is widely regarded as a potential communication paradigm that holds the promise of breaking through the ‘Shannon trap’ \cite{yang2022semantic}. It works on the principle of semantic transmission, that is, the semantics represented by the transmitted bits. The meaning or characteristics of the message being sent are first extracted at the sender, and the semantic information is obtained at the receiver through a matched knowledge base (KB) between the transmitter and receiver \cite{luo2022semantic}. Semantic communication focuses on whether the meaning of the transmitted information is received by the receiver rather than the accurate stream of bits transmitted \cite{xie2021deep}. Therefore, there may be a syntactic mismatch in SemCom, but the semantics of the translation is correct. In other words, even if the order or phase of the utterances received at the receiver changes, the semantic is still correctly understood. The difference between SemCom and ordinary communication is that ordinary communication is ‘transmit-before-understand’, while semantic communication is ‘understand- before-transmit. With the support of AI techniques, end devices can have human-like inference capabilities so that SemCom can be achieved by integrating semantic extraction into the communication model, where only interesting information of the receiver will be transmitted instead of the raw data, which not only alleviates the pressure of bandwidth but also strengthens data privacy by reducing the exchange of redundant data.

Semantic communication has some promising scenarios, such as being used in IoT networks, smart vehicle networks and smart Factory \cite{luo2022semantic}. In IoT networks, various intelligent devices such as UAVs and sensors are used to perform monitoring and tracking tasks which require them to have intelligent detection, communication, and data processing capabilities. For example, UAV swarms are used to monitor large areas of forests for fire prevention, where large amounts of images, video and audio are collected by UAVs and a large amount of data needs to be uploaded to the cloud or mobile edge computing (MEC) servers for processing. Using semantic communication can greatly save communication resources and reduce upload delay. This enables UAVs to respond more quickly in emergency situations. Because it only transmits the fire related semantic information. In \cite{8723589}\cite{xie2021deep}, the average running time of the semantic communication system is compared with that of traditional schemes. The results show that deep learning-based SemCom can help to reduce latency by compressing and extracting semantic information. Therefore, semantic communication is suitable for the transmission of large amounts of data e.g., for smart vehicles, and it is more resistant to channel noise and interference than the bit streams transmitted in conventional communication, enhancing the dependability of data transmission. 

However, training the semantic extraction model requires a lot of computing and storage resources, which is difficult for edge devices to implement sophisticated DNNs on board. DNN model compression techniques (quantification and network sparsification) and distributed learning are effective ways to address the above issue\cite{luo2022semantic, xie2020lite}. Specifically, in UAV swarms, FL can be applied to train the deep learning-based semantic communication model. Each UAV can participate in the DNN training processes by using its locally sensed data, which not only saves a lot of computing power for each device but also speed up the training process. This shows that SemCom-UAVs based on distributed learning can train models faster and reduce communication latency by transmitting only the semantic information of interest, as a promising approach that can be applied to more UAV swarm tasks in the future. Fig.~\ref{fig:RIS} illustrates two types of UAV swarm-semantic applications, one is that UAV swarms are used to enhance semantic transmission, where semantic noise and Gaussian noise are considered to ensure semantic correctness during transmission. In another way, UAV swarms can be used to perform semantic encoding and channel encoding  at the user side as well as channel decoding and semantic decoding at the base station side.

In \cite{xie2020lite}, a distributed SemCom system L-DeepSC was proposed for IoT networks to alleviate the burden on IoT devices with limited computing resources and power by reducing the resolution of weights and pruning the redundancy of model, where the task of L-DeepSC is to transmit low-complexity text by semantic communication between IoT devices and cloud or edge servers. A channel state information (CSI) assisted training processing method is also proposed to reduce the impact of fading channels on transmission. Simulation results show that the proposed DeepSC significantly outperforms conventional methods in low SNR and can achieve compression ratios of up to 40x without performance degradation. which is regarded as a promising candidate for future smart IoT.

In \cite{yun2021attention}, in order to provide ultra-reliable and low-latency communication (URLLC) to ground mobile users, a novel centralized training and decentralized execution (CTDE) MARL framework named a graph attention exchange network (GAXNet) for UAV aided non-terrestrial URLLC was proposed, where the URLLC problem is done by real-time control of multiple UAVs and collision avoidance between UAVs. In GAXNet, an attention graph is constructed for each UAV, which locally measures its attention level to neighboring UAVs, and simultaneously exchanges semantically meaningful attention weights with other UAVs to minimize the attention mismatch between them. The simulation results show that the proposed approach GAXNet has lower latency and error rates compared to the state-of-the-art CTDE approach QMIX, improving the reliability of air-to-ground networks.

\begin{figure*}[!h]
  \centering
    \subfigure[UAV swarms to enhance semantic information transmission. ]{\includegraphics[width=0.9\textwidth]{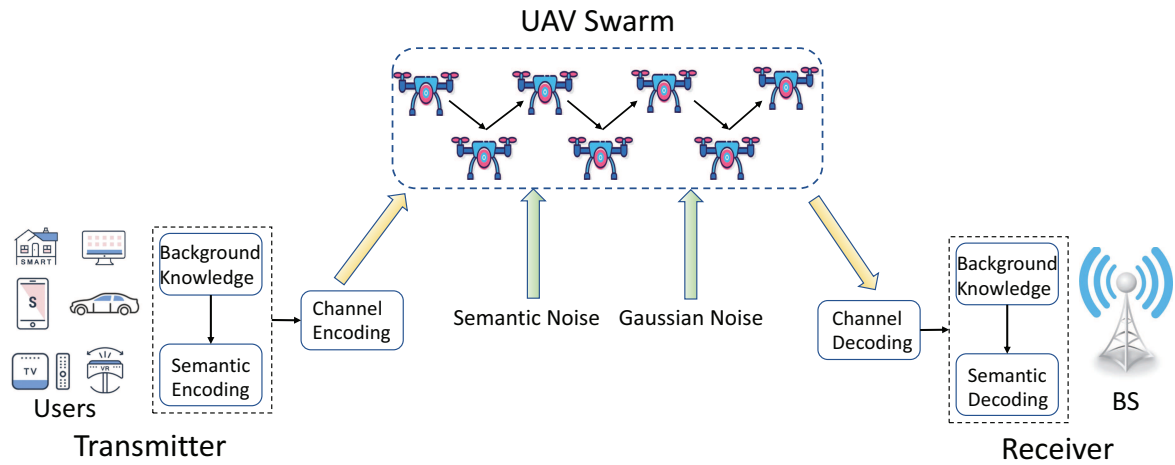}} 
    \subfigure[UAV swarms for semantic communication encoding and decoding.  ]{\includegraphics[width=0.9\textwidth]{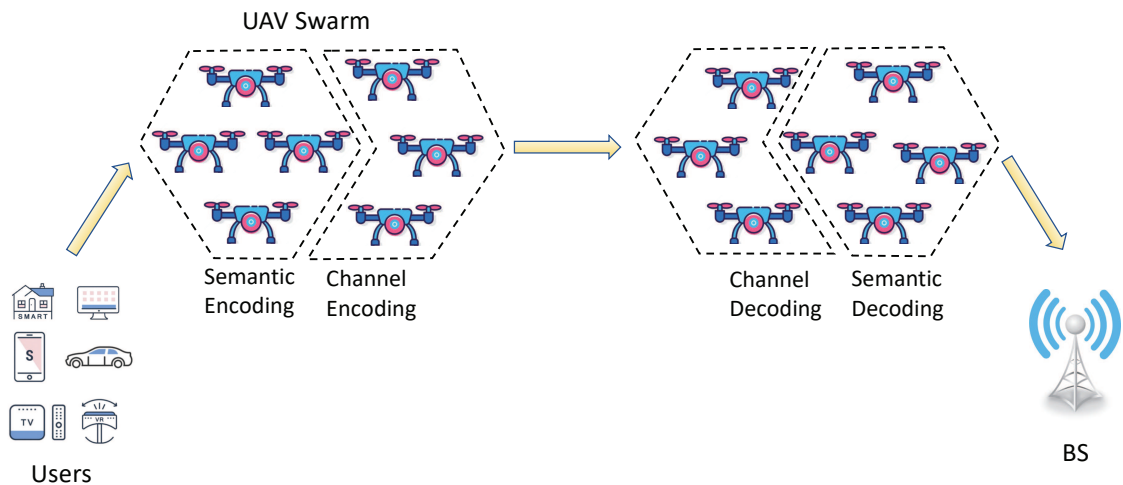}} 
    
  \caption{Applications of UAV swarm-enabled RIS}
	\label{fig:RIS}
\end{figure*}

\subsection{UAV Swarms for Virtual Reality (VR)}

\begin{figure}[!htbp]
    \centering
    \includegraphics[width=0.5\textwidth]{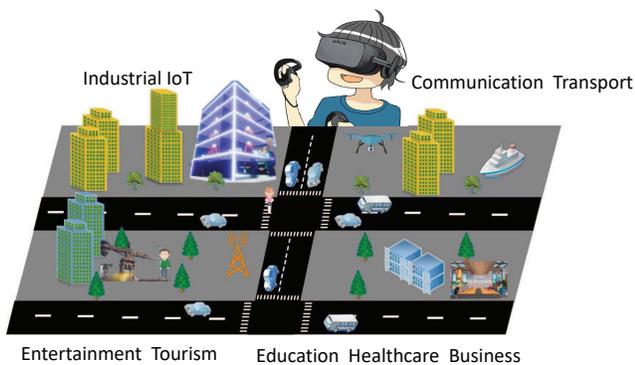}
    \caption{Four representative applications of using VR in metaverse.}
    \label{fig:VR}
\end{figure}

A new revolution of IoT is the metaverse, which is created through the integration and interplay of multiple emerging technologies such as AI, big data, blockchain, VR, etc. In the following, we briefly introduce the roles that UAVs can play in the metaverse services and VR applications.

VR is an immersive technology that allows users to interact with characters in a virtual environment created entirely by a computer instead of a real scene.  The development of VR technology is set to revolutionize the way people communicate\cite{zong20196g}, and it has a wide range of application scenarios.
As an emerging application in future wireless communications, metaverse can be used as an integration of physical society and digital society, where VR can play an important role. 
In metaverse, VR based services can be used in education, tourism, entertainment, healthcare, business and IoT, as shown in Fig.~\ref{fig:VR}. In particular, the covid-19 in recent years has led to many schools going online, and VR technology has made online classes more interesting for students as if they were sitting in a classroom. VR technology can also simulate ancient scenes, allowing us to travel with a clearer understanding of the history of the time, and used in medical surgery simulations, allowing doctors to become more skilled at the surgery. In general, VR devices are tiny and have limited computing resources, leading to the proposed combination of drone and VR technology. Due to their flexibility and stealthiness, UAVs can act as edge servers or relays to collect, process and transmit VR content. For example, in \cite{8736996}, they investigate UAV-IoT networks to implement VR in remote scenes, where multiple UAVs are distributed at locations of interest, collecting data from different viewpoints, transmitting them to the aggregation point, and receiving control information transmitted. The aggregator then proceeds to construct a VR representation for the user. They used VR immersion fidelity as a function of UAV-IoT capture rate and maximized fidelity by studying fast RL to find the optimal dynamic UAV location over interesting scenes.  The work in \cite{zhou2020communication} presented a UAV based station (U-BS) based low-latency VR delivery system, where the task of the UAV is to transmit VR content from the cloud server to multiple terrestrial VR users and also caches the popular VR input data to mitigate the backhaul latency further. They jointly considered UAVs' location, communication, computing capacity allocation, and caching policy and proposed an effective iterative algorithm approach to tackle the above non-convex optimization problems. In addition, multi-UAVs are considered to solve the problem of limited battery capacity. 

As is well known, rendering is the critical bottleneck in wireless VR systems. In \cite{guo2020adaptive}, the authors consider an adaptive VR framework based on mmWave and MEC, where real-time VR rendering can be offloaded to MEC. The framework aims to maximize user quality of experience (QoE) by jointly optimizing the rendering offload pattern and caching policy. A distributed learning based on offline training and online running algorithm is proposed to solve the above optimization problem, where DRL is used for offline training and game theory is used for online running. Simulation results show that the proposed algorithm is superior in terms of scalability and adaptability. UAVs can be used as MEC servers to provide caching and rendering services to VR users. Similarly, the authors in \cite{ding2021uav} propose a DRL-based UAV approach to assist the rendering process of VR devices, where the drone acts as an aerial MEC server to supply real-time rendering of random VR tasks. Both the trajectory of the UAV and the VR rendering modes are jointly optimized to achieve a maximum rendering completion rate. This problem is modelled as a Markov decision process (MDP) and the Twin Delayed Deep Deterministic Policy Gradient (TD3) algorithm is used in the DRL framework to find the optimal strategy for the UAV trajectory and VR rendering mode. Simulation results demonstrated that the proposed approach could allocate computational resources among VR devices and the UAV reasonably and is significantly better than the DDPG-based baseline algorithm in terms of convergence speed and rendering completion rate. However, this paper focuses on a single UAV and can be extended to UAV swarm scenarios in the future.

\section{Conclusion and Open Discussion}\label{con}
In this paper, we have introduced four DL algorithms that can be used for UAV swarms. The principle, advantages and disadvantages of each DL algorithm are discussed. We further provided the various optimization problem cases in the different scenarios of using DL for UAV swarms. In particular, the performance metrics about communication, computation, sensing and caching are analyzed. The emerging applications with UAV swarms are presented in the pointview of utilizing DL.


With the rapid development of DL and UAV communication, there are still many future directions including the architecture, efficient design, and experiments. 
\begin{itemize}
\item {\bf{Architecture and protocol design}}: The native  network architectures and protocols for DL in UAV swarms should consider both the unique wireless properties of wireless channels and the advantages of  swarm intelligence. 
One promising architecture is to fully utilize the distributed semantic communication framework, which can ensure limited and efficient wireless transmission among UAVs and well adapt to task-oriented transmissions. Since DL involves the model and dataset exchange, the centralized and distributed data storage with DL for UAV swarms should be taken into consideration for the network architectures and protocols design.  Further considering the labelling efforts of the raw data by UAV swarms, wireless crowd labelling is an important direction. 
 \item {\bf{Efficient communication and computation design}}: Due to coupled communication and computation resources and large number of UAVs, it is of importance to investigate the decentralized transmission optimization for DL in UAV swarms. Considering massive number of UAVs, over-the-air computation is an efficient technique to solve the massive access issues for DL in UAV swarms. 
 To further guarantee the real-time and high-mobility of UAVs, the ultra-low latency edge learning and distributed inference in UAV swarms is an important direction. 
 As a result, the joint communication and computation considering calls for the energy efficient schemes for DL in UAV swarms.
 \item {\bf{Experiments and testbeds using the swarm intelligence}}: To verify the effectiveness of DL in UAV swarms, many aspects should be considered in the real scenarios, which include wireless communication, intelligent control, machine learning, and dynamics. Thus, the real experiments and testbeds using the swarm intelligence are important. 
\end{itemize}

\bibliographystyle{IEEEtran}
\bibliography{ref}




%








\begin{IEEEbiographynophoto}{Yahao Ding}
	(Student Member, IEEE) received the Master degree in Communications and Signal Processing from Imperial College London, UK in 2020. She is currently pursuing the PhD degree in information and communication engineering with King's College London. Her current research interests include federated learning, security, and UAV swarms.
\end{IEEEbiographynophoto}

\begin{IEEEbiographynophoto}{Zhaohui~Yang} (Member, IEEE) received the B.S. degree in information science and engineering from the Chien-Shiung Wu Honors College, Southeast
University, Nanjing, China, in June 2014, and the Ph.D. degree in communication and information system from the National Mobile Communications Research Laboratory, Southeast University, in May 2018. From May 2018 to October 2020, he was a Post-Doctoral Research Associate with the Center for Telecommunications Research, Department of Informatics, King’s College London, U.K.
He is currently a Visiting Associate Professor with the Zhejiang Key Laboratory of Information Processing Communication and Networking, College of Information Science and Electronic Engineering, Zhejiang University, and a Research Fellow with the Department of Electronic and Electrical Engineering, University College London, U.K. His research interests include federated
learning, reconfigurable intelligent surface, UAV, and NOMA. He was a TPC Member of IEEE ICC from 2015 to 2021 and GLOBECOM from 2017 to 2021. He was the Co-Chair of workshops on edge learning and wireless communications in several conferences, including the IEEE International Conference on Communications (ICC), the IEEE Global Telecommunication Conference (GLOBECOM), the IEEE Wireless Communications and Networking Conference (WCNC), and the IEEE International Symposium on Personal, Indoor and Mobile Radio Communication (PIMRC). He was an Exemplary Reviewer of IEEE TRANSACTIONS ON COMMUNICATIONS in 2019 and 2020. He has guest edited a feature topic of IEEE Communications Magazine on Communication Technologies for Efficient Edge Learning. He is also an Associate Editor of IEEE COMMUNICATIONS LETTERS, IET
Communications, and EURASIP Journal on Wireless Communications and Networking.
\end{IEEEbiographynophoto}

\begin{IEEEbiographynophoto}{Quoc-Viet~Pham} (Member, IEEE)  received the B.S. degree in electronics and telecommunications engineering from the Hanoi University of Science and
Technology, Hanoi, Vietnam, in 2013, and the M.S. and Ph.D. degrees in telecommunications engineering from Inje University, Gimhae, South Korea, in 2015 and 2017, respectively.
From September 2017 to December 2019, he was with Kyung Hee University, Seoul, South Korea, Changwon National University, Changwon, South Korea, and Inje University, Gimhae, South Korea,
on various academic positions. He is currently a Research Professor with the Korean Southeast Center for the 4th Industrial Revolution Leader Education, Pusan National University, Busan, South Korea. He has been granted the
Korea NRF Funding for outstanding young researchers for the term 2019– 2023. His research interests include convex optimization, game theory, and machine learning to analyze and optimize edge/cloud computing systems and beyond 5G networks.
Dr. Pham received the Best Ph.D. Dissertation Award in Engineering from Inje University, in 2017, and the top reviewer award from the IEEE TRANSACTIONS ON VEHICULAR TECHNOLOGY, in 2020. He is an Editor Journal of Network and Computer Applications (Elsevier).
\end{IEEEbiographynophoto}

\begin{IEEEbiographynophoto}{Zhaoyang~Zhang} (Senior Member, IEEE) received the Ph.D. degree from Zhejiang University, Hangzhou, China, in 1998. He is currently a Qiushi Distinguished Professor with Zhejiang University. He has coauthored more
than 300 peer-reviewed international journal and conference papers. His current research interests include fundamental aspects of wireless communications and networking, such as information theory and coding, network signal processing and distributed learning, AI-empowered communications and networking, network intelligence with synergetic sensing, computing and communication, and so on. He was a co-recipient of eight best paper awards of international conferences, including IEEE ICC 2019 and IEEE GlobeCom 2020. He was awarded the National Natural Science Fund for Distinguished Young Scholars from NSFC in 2017. He is serving or has served as an Editor for IEEE TRANSACTIONS ON WIRELESS COMMUNICATIONS, IEEE TRANSACTIONS ON COMMUNICATIONS, and IET Communications, and as a General Chair, TPC Co-Chair, or Symposium Co-Chair for VTC-Spring
2017 Workshop HMWC, WCSP 2013/2018, and Globecom 2014 Wireless
Communications Symposium. He was also a Keynote Speaker for APCC
2018 and VTC-Fall 2017 Workshop NOMA
\end{IEEEbiographynophoto}

\begin{IEEEbiographynophoto}{Mohammad Shikh-Bahaei}
(Senior Member, IEEE) received the B.Sc. degree from the University of Tehran, Tehran, Iran, in 1992, the M.Sc. degree
from the Sharif University of Technology, Tehran, in 1994, and the Ph.D. degree from King’s College London, U.K., in 2000. He worked for two start-up companies, and National Semiconductor
Corporation, Santa Clara, CA, USA (now part of Texas Instruments Inc.), on the design of third-generation (3G) mobile handsets, for which he has been awarded three U.S. patents as Inventor and
Co-Inventor, respectively. In 2002, he joined King’s College London, as a Lecturer, where he is currently a Full Professor in telecommunications with the Center for Telecommunications Research, Department of Engineering. He has authored or coauthored numerous journals and conference papers. He has been engaged in research in the area of wireless communications and
signal processing for 25 years both in academic and industrial organizations. His research interests include learning-based resource allocation for multimedia applications over heterogeneous communication networks, full duplex and UAV communications, and secure communication over wireless networks. He was a recipient of the overall King’s College London
Excellence in Supervisory Award in 2014. He was the Founder and the Chair of the Wireless Advanced (formerly, SPWC) Annual International Conference from 2003 to 2012
\end{IEEEbiographynophoto}

\end{document}